\documentclass[conference]{IEEEtran}

\IEEEoverridecommandlockouts
\usepackage{cite}
\usepackage{amsmath,amssymb,amsfonts}
\usepackage{algorithmic}
\usepackage{graphicx}
\usepackage{textcomp}
\usepackage{xcolor}
\usepackage{url}
\usepackage{courier}
\usepackage{graphicx}
\usepackage{balance}  
\usepackage{blindtext}
\usepackage[ruled]{algorithm2e}
\usepackage{booktabs}
\usepackage{multirow}
\usepackage{amsfonts}
\usepackage{diagbox}
\usepackage{CJK}
\usepackage{amsmath}
\usepackage{enumitem}
\usepackage{xcolor}
\usepackage{boxedminipage}
\usepackage{subfigure}
\usepackage{array}
\newcolumntype{b}{>{\global\let\currentrowstyle\relax}}  
\newcolumntype{t}{>{\currentrowstyle}}  
\newcommand{\rowstyle}[1]{%
  \gdef\currentrowstyle{#1}%
  #1\ignorespaces  
} 

\def\BibTeX{{\rm B\kern-.05em{\sc i\kern-.025em b}\kern-.08em
    T\kern-.1667em\lower.7ex\hbox{E}\kern-.125emX}}
\begin{document}

\title{Self-paced Ensemble for Highly Imbalanced Massive Data Classification
\thanks{
    This work was conducted when the first author was an intern at Microsoft Research Asia.
    This work is partially supported by National Natural Science Foundation of China (No.61976102).
    }
}

\author{
\IEEEauthorblockN{
  Zhining Liu\IEEEauthorrefmark{1}\IEEEauthorrefmark{2}, 
  Wei Cao\IEEEauthorrefmark{3}, 
  Zhifeng Gao\IEEEauthorrefmark{3}, 
  Jiang Bian\IEEEauthorrefmark{3}, 
  Hechang Chen\IEEEauthorrefmark{1}\IEEEauthorrefmark{2}, 
  Yi Chang\IEEEauthorrefmark{1}\IEEEauthorrefmark{2} and 
  Tie-Yan Liu\IEEEauthorrefmark{3}}
\IEEEauthorblockA{
  \IEEEauthorrefmark{1}{\it School of Artificial Intelligence, Jilin University}\\
  \IEEEauthorrefmark{2}{\it Key Lab. of Symbolic Computation and Knowledge Engineering of MOE, Jilin University}\\
  \IEEEauthorrefmark{3}{\it Microsoft Research}\\
  znliu19@mails.jlu.edu.cn, \\
  \{weicao, zhgao, jiang.bian, tyliu\}@microsoft.com,\\
  chenhc14@mails.jlu.edu.cn, yichang@jlu.edu.cn
}
}

\maketitle

\begin{abstract}
Many real-world applications reveal difficulties in learning classifiers from imbalanced data. The rising big data era has been witnessing more classification tasks with large-scale but extremely imbalance and low-quality datasets. Most of existing learning methods suffer from poor performance or low computation efficiency under such a scenario. To tackle this problem, we conduct deep investigations into the nature of class imbalance, which reveals that not only the disproportion between classes, but also other difficulties embedded in the nature of data, especially, noises and class overlapping, prevent us from learning effective classifiers. Taking those factors into consideration, we propose a novel framework for imbalance classification that aims to generate a strong ensemble by self-paced harmonizing data hardness via under-sampling. Extensive experiments have shown that this new framework, while being very computationally efficient, can lead to robust performance even under highly overlapping classes and extremely skewed distribution. Note that, our methods can be easily adapted to most of existing learning methods (e.g., C4.5, SVM, GBDT and Neural Network) to boost their performance on imbalanced data. 
\end{abstract}


\section{Introduction}

The development of information technology brings the explosion of massive data in our daily life. However, many real applications usually generate very imbalanced datasets for corresponding key classification tasks. 
For instance, online advertising services can give rise to a high amount of datasets, consisting of user views or clicks on ads, for the task of click-through rate prediction \cite{graepel2010ctr}. Commonly, user clicks only constitute a small rate of user behaviors .
For another example, credit fraud detection \cite{dal2018creditfraud} relies on the dataset containing massive real credit card transactions where only a small proportion are frauds. 
Similar situations also exist in the tasks of medical diagnosis, record linkage and network intrusion detection etc \cite{gamberger1999medical,sariyar2011record-linkage,haixiang2017overview}.
In addition, real-world datasets are likely to contain other difficulty factors, including noises and missing values.
Such {\em highly imbalanced}, {\em large-scale} and {\em noisy} data brings serious challenges of downstream classification tasks.

Traditional classification algorithms (e.g., C4.5, SVM or Neural Networks \cite{quinlan1986dt,cortes1995svm,haykin2009neuralnetworks}) demonstrate unsatisfactory performance on imbalanced datasets. 
The situation can be even worse when the dataset is large-scale and noisy at the same time.
Attribute to their inappropriate presuming on relatively balanced distribution between positive and negative samples, the minority class is usually ignored due to the overwhelming number of majority instances. On the other hand, the minority class usually carries the concepts with greater interests than majority class \cite{he2008overview, he2013overview}. 

To overcome such issue, a series of research work has been proposed, which can be classified into three categories:
\begin{itemize}[leftmargin=0.14in]
    \setlength{\itemsep}{0pt}
    \item {\em Data-level} methods modify the collection of examples to balanced distributions and / or remove difficult samples. They may be inapplicable on datasets with categorical features or missing values due to their distance-based design (e.g., NearMiss, Tomeklink \cite{mani2003nearmiss,tomek1976tomeklink}). Besides, they suffer from large computational cost (e.g., SMOTE, ADASYN \cite{chawla2002smote,he2008adasyn}) when applying on large-scale data.
    \item {\em Algorithm-level} methods directly modify existing learning algorithms to alleviate the bias towards majority objects.
    However, they require assistance from domain experts before-hand (e.g., setting cost matrix in cost-sensitive learning~\cite{elkan2001cost-sensitive,liu2006cost-sensitive-imbalance}). They may also fail when cooperating with batch-training classifiers like neural network since they do not balance the class distribution on the training data.
    \item {\em Ensemble} methods combine one of the previous approaches with an ensemble learning algorithm to form an ensemble classifier. Some of them suffer from large training cost and poor applicability (e.g., SMOTEBagging \cite{wang2009smotebagging}) on realistic tasks. The other ones potentially lead to underfitting or overfitting (e.g., EasyEnsemble, BalanceCascade \cite{liu2009ee-bc}) when the dataset is highly noisy.
\end{itemize}

For above reasons and more, none of the prevailing methods can well handle the {\em highly imbalanced}, {\em large-scale} and {\em noisy} classification task, while it is a common problem in real-world applications.
The main reason behind existing methods' failure on such tasks is that they ignored difficulties embedded in the nature of imbalance learning. 
Not only the class imbalance itself, other factors like presence of noise samples \cite{napierala2010learn-from-noisy-data} and overlapped underlying distribution between the classes \cite{garcia2007overlap,prati2004overlap-small-disjuncts} also significantly deteriorate the classification performance. 
Their influences can be further enlarged by the high imbalance ratio. 
Besides, different models show various sensitivity to these factors.
For above reasons, all these factors need to be considered to achieve more accurate classification.

We introduce the concept of ``{\em classification hardness}'' to integrate aforementioned difficulties.
Intuitively, hardness represents the difficulty of correctly classifying a sample for a specific classifier. 
Thus the distribution of classification hardness implicitly contains the information of task difficulties. 
For example, noises are likely to have large hardness values and the proportion of high-hardness samples reflected the level of class overlapping. 
Moreover, hardness distribution is naturally adaptive to different models since it was defined with respect to given classifier.
Such hardness distribution can be used to guide the re-sampling strategy to achieve better performance.

\begin{figure}[tbh]
    \centering
    \includegraphics[width=1\linewidth]{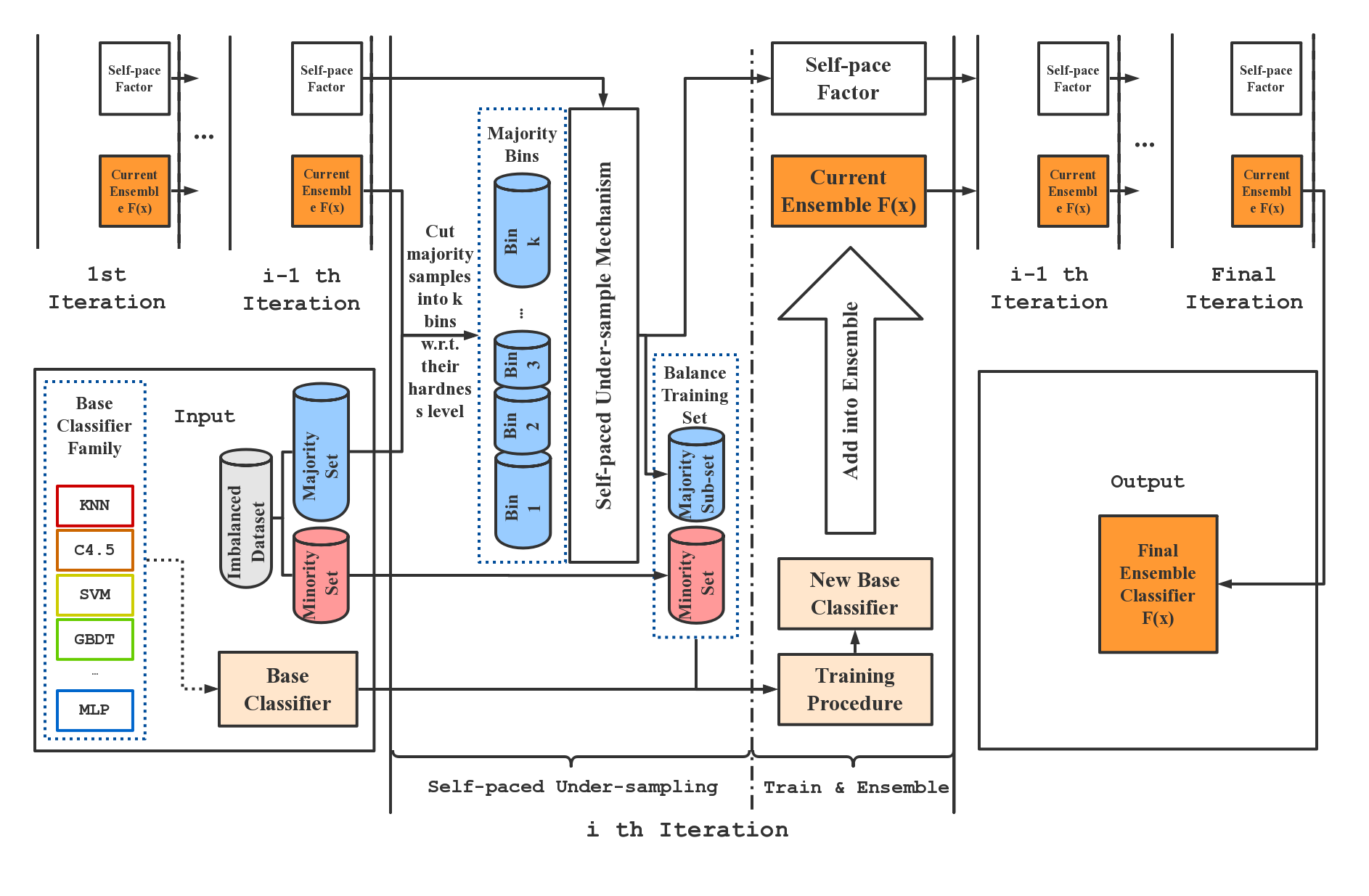}
    \caption{Self-Paced Ensemble Process.}
    \label{fig:process}
\end{figure}

Based on the classification hardness, we propose a novel learning framework called {\em Self-paced Ensemble} (abbreviated as $\texttt{SPE}$) in this paper. Instead of simply balancing the positive/negative data or directly assigning instance weights, we consider the distribution of classification hardness over the dataset, and iteratively select the most informative majority data samples according to the hardness distribution. The under-sampling strategy is controlled by a self-paced procedure. Such self-paced procedure enables our framework gradually focuses on the harder data samples, while still keeps the knowledge of easy sample distribution in order to prevent overfitting.
Fig. \ref{fig:process} shows the pipeline of self-paced ensemble. 

In summary, the contributions of this paper are as follows:
\begin{itemize}[leftmargin=0.14in]
    \item In this paper we demonstrate the reason of conventional imbalance learning methods failing on the real-world massive imbalanced classification task. We conduct comprehensive experiments with analysis and visualization that can be valuable for other similar classification systems. 
    \item We proposed {\em Self-paced Ensemble} ($\texttt{SPE}$), a learning framework for massive imbalanced data classification. $\texttt{SPE}$ can be used to boost any canonical classifier's performance 
    (e.g., C4.5, SVM, GBDT, and Neural Network) 
    on real-world highly imbalanced tasks while being very computationally efficient. Comparing with the existing methods, $\texttt{SPE}$ is accurate, fast, robust, and adaptive.
    \item We introduce the concept of classification hardness. By considering the distribution of classification hardness over the dataset, the learning procedure of our proposed framework $\texttt{SPE}$ is automatically optimized in a model-specific way. Unlike prevailing methods, our learning framework does not require any pre-defined distance metrics which is usually unavailable in real-world scenarios.
\end{itemize}

\section{Problem definition}

In this section, we first describe the class imbalance problem considered in this paper. Then we give some necessary symbol definition and show the evaluation criteria that are commonly used in imbalanced scenarios.

{\em Class imbalance:} A dataset is said to be imbalanced whenever the number of instances from the different classes is not nearly the same. Class imbalance exists in the nature of various real-world applications, like medicine (sick vs. healthy), fraud detection (normal vs. fraud), or click-through-rate prediction (clicked vs. ignored). The uneven distribution poses a difficulty for applying canonical learning algorithms on imbalanced dataset, as they will be biased towards the majority group due to their accuracy-oriented design. 
Despite such problem has been extensively studied, in real applications, class imbalance often co-exists with other difficulty factors, such as enormous data scale, noises, and missing values. Therefore, the performances of existing methods are still unsatisfactory.

{\em Symbol definition:} In this paper, we only consider the binary situation that exists widely in practical applications \cite{dal2018creditfraud,he2008overview,he2013overview}. In binary imbalance classification, only two classes were considered: the {\em minority} class with less samples and the {\em majority} class with relatively more samples. For simplicity, in this paper we always let the minority class to be positive class and the majority class to be negative. We use $\mathcal{D}$ to denote the collection of all training samples $(x, y)$. The minority class set $\mathcal{P}$ and majority class set $\mathcal{N}$ are then defined as:
$$\mathcal{P}=\{(x, y)\ |\ y=1\},\ \mathcal{N}=\{(x, y)\ |\ y=0\}$$

Therefore, we have $|\mathcal{N}|\gg|\mathcal{P}|$ for (highly) imbalanced problems. In order to uniformly describe the level of class imbalance in different datasets, we consider the \underline{Imbalance Ratio (IR)}, which is defined as the number of majority class examples divided by the number of minority class examples:
$$\text{\em Imbalanced Ratio (IR)}=\frac{n_{majority}}{n_{minority}}= \frac{|\mathcal{N}|}{|\mathcal{P}|}$$

\vspace{2mm}

{\em Evaluation criteria:} Since the accuracy does not well reflect the model performance, we usually adopt the other evaluation criteria based on the number of true / false positive / negative prediction. Under the binary scenario, the results of the correctly and incorrectly recognized examples of each class can be recorded in a confusion matrix. Table \ref{confusion-matrix} shows the confusion matrix for binary classification.

\begin{table}[h]
\small
\renewcommand{\arraystretch}{1.1}
\caption{Confusion matrix for binary classification.}
\label{confusion-matrix}
\centering
\begin{tabular}{|c|cc|}
\hline
\renewcommand{\arraystretch}{1.1}
\rowstyle{\bfseries}
\diagbox[width=6.5em]{Label}{Predict} & Positive & Negative\\
\hline
Positive & True Positive (TP) & False Negative (FN)\\
Negative & False Positive (FP) & True Negative (TN)\\
\hline
\end{tabular}
\end{table}

For evaluating the performance on minority class, recall and precision are commonly used. Furthermore, we also consider the F1-score, G-mean (i.e., harmonic / geometric mean of precision and recall)~\cite{powers2011evaluation,sokolova2006evaluation}, MCC (i.e., Matthews correlation coefficient)~\cite{boughorbel2017mcc}, and AUCPRC (i.e., the area under the precision-recall curve)~\cite{davis2006aucprc}.
\begin{itemize}[leftmargin=0.14in, label=-]
    \item $Recall = \frac{TP}{TP+FN}$
    \item $Precision = \frac{TP}{TP+FP}$
    \item {\em F1-score} = $2\cdot\frac{Recall\times Precision}{Recall+Precision}$
    \item {\em G-mean} = $\sqrt{Recall \cdot Precision}$
    \item {\em MCC} = $\frac{TP\times TN-FP\times FN}{\sqrt{(TP+FP)(TP+FN)(TN+FP)(TN+FN)}}$
    \item {\em AUCPRC} = {\em Area Under Precision-Recall Curve}
\end{itemize}

\section{Limitations of Existing Methods}

In this section, we give a brief introduction to existing imbalance learning solutions, and discuss why they obtain unsatisfactory performance on the real-world industrial tasks.
To solve the class imbalance problem, researchers have proposed a variety of methods. This research field is also known as {\em imbalance learning}. As mentioned in the introduction, we categorize existing imbalance learning methods into three groups: {\em Data-level}, {\em Algorithm-level} and {\em Ensemble}.

{\em Data-level Methods}: This group of methods concentrates on modifying the training set to make it suitable for a standard learning algorithm. With respect to balancing distributions, data-level methods can be categorized into three groups:
\begin{itemize}[leftmargin=0.14in]
    \item {Under-sampling} approaches that remove samples from the majority class (e.g., \cite{kubat1997oss,laurikkala2001ncr,tomek1976tomeklink}).
    \item {Over-sampling} approaches that generate new objects for the minority class (e.g., \cite{chawla2002smote,han2005borderline-smote,he2008adasyn}).
    \item {Hybrid-sampling} approaches that combine two methods above (e.g., \cite{batista2003smotetomek,batista2004smoteenn}).
\end{itemize}

Standard random re-sampling methods often lead to removal of important samples or introduction of meaningless new objects. Therefore, more advanced methods were proposed that try to maintain structures of groups and/or generate new data according to underlying distributions. They apply k-Nearest Neighborhood (k-NN) algorithm \cite{altman1992knn} to extract underlying distribution in the feature space, and use that information to guide their re-sampling.

However, the application of k-NN algorithm requires pre-defined distance metric, which is usually unavailable in the real-world datasets since they may contain categorical features and/or missing values. 
k-NN algorithm is also easily disturbed by noises thus unable to reveal the underlying distribution for re-sampling methods when the dataset is noisy. 
Moreover, the computational cost of applying k-NN grows quadratically with the size of the dataset. Thus running such distance-based re-sampling methods on large-scale datasets can be extremely slow.

\begin{figure*}[h!]
    \centering
    \subfigure[Non-overlapped Dataset]{
        \label{fig:growimb_nop}
        \includegraphics[width=0.29\linewidth]{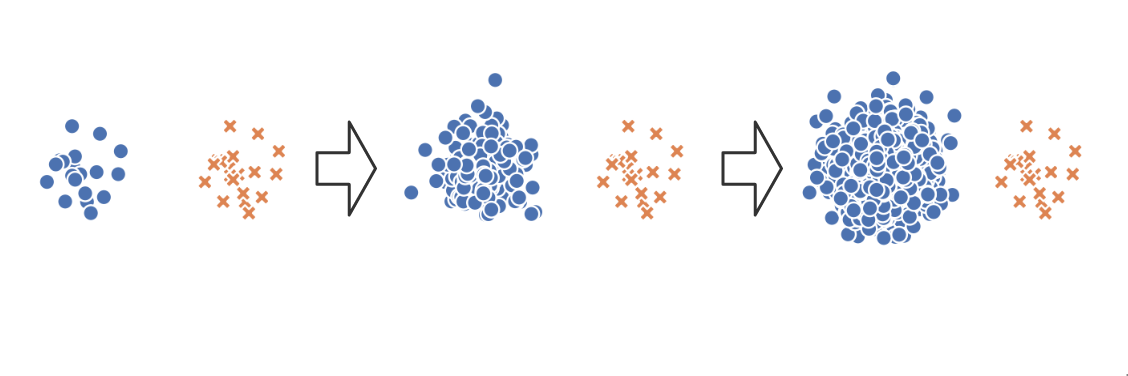}
    }
    \subfigure[Hardness (KNN~\cite{altman1992knn})]{
        \label{fig:growimb_nop_knn}
        \includegraphics[width=0.29\linewidth]{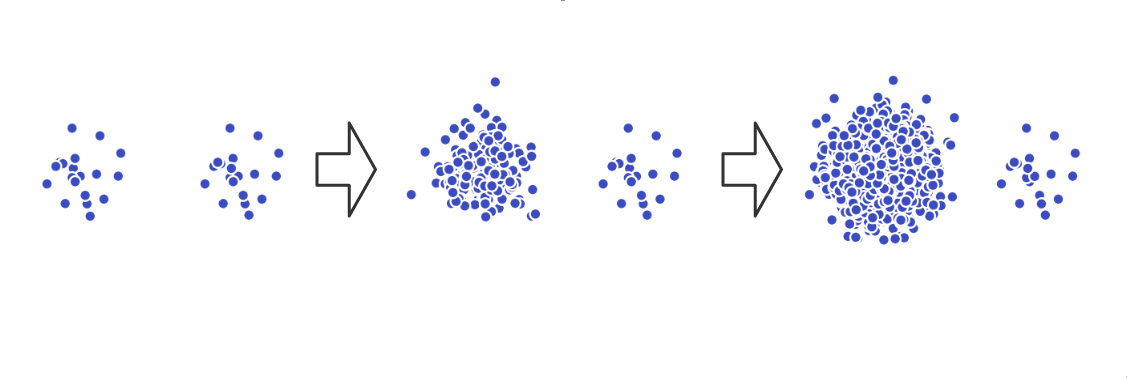}
    }
    \subfigure[Hardness (AdaBoost~\cite{freund1997adaboost})]{
        \label{fig:growimb_nop_ada}
        \includegraphics[width=0.29\linewidth]{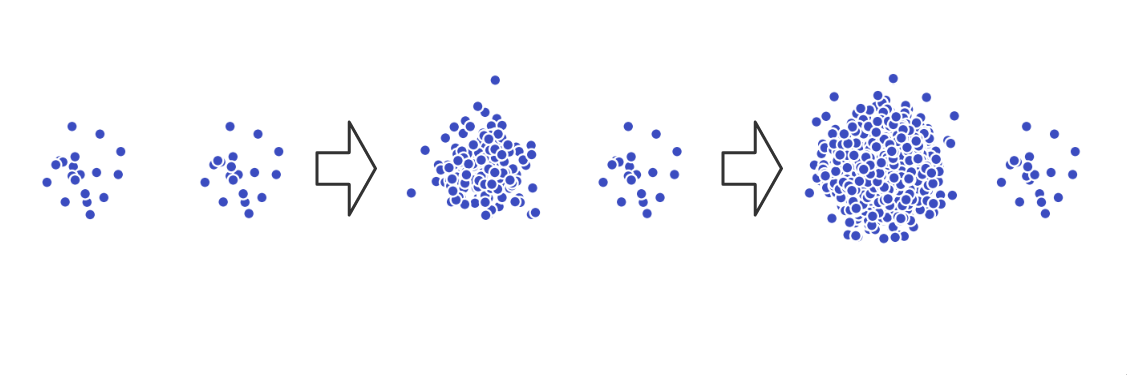}
    }
    \\
    \subfigure[Overlapped Dataset]{
        \label{fig:growimb_op}
        \includegraphics[width=0.29\linewidth]{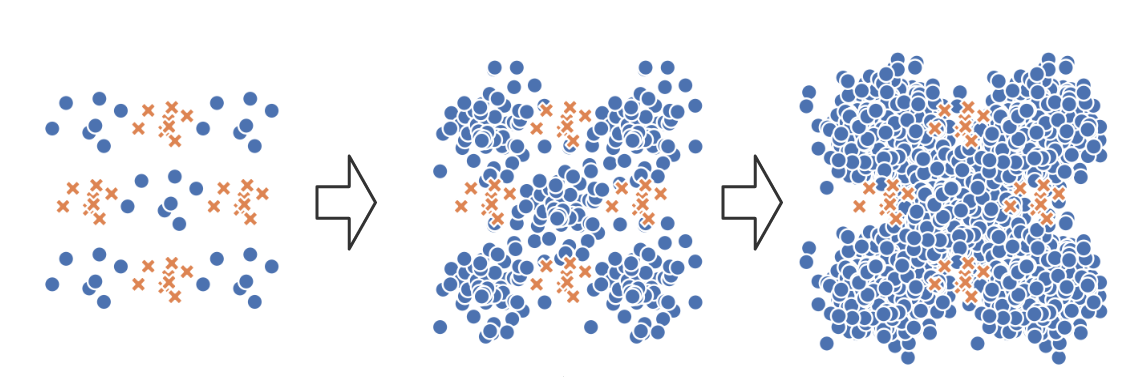}
    }
    \subfigure[Hardness (KNN~\cite{altman1992knn})]{
        \label{fig:growimb_op_knn}
        \includegraphics[width=0.29\linewidth]{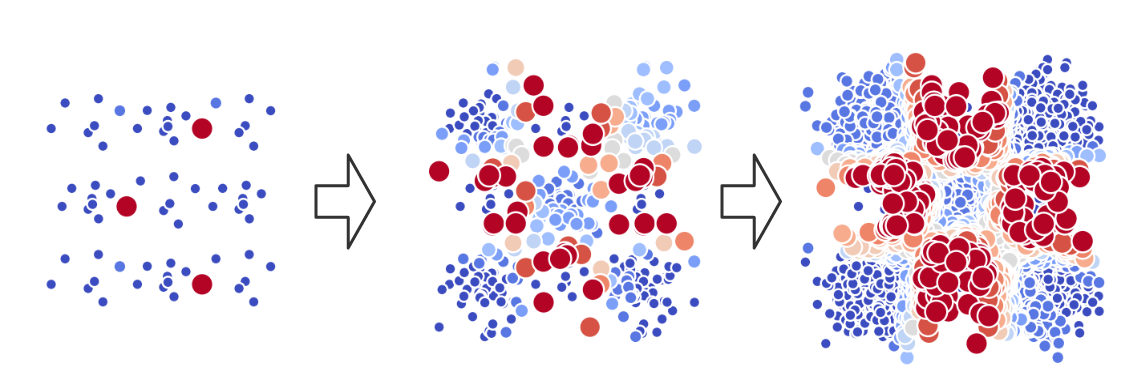}
    }
    \subfigure[Hardness (AdaBoost~\cite{freund1997adaboost})]{
        \label{fig:growimb_op_ada}
        \includegraphics[width=0.29\linewidth]{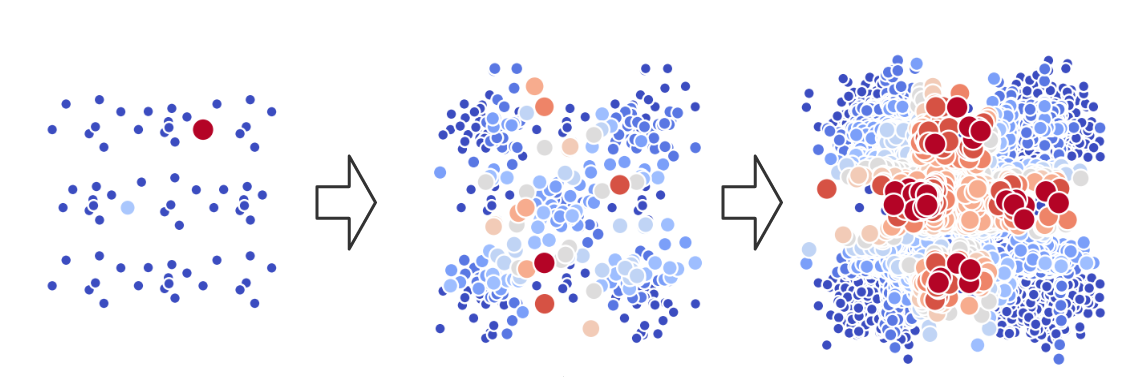}
    }
    \caption{Comparison of overlapped / non-overlapped dataset under different level of class imbalance. (a)(c) shows the original datasets, (b)(e) are the hardness w.r.t. KNN classifier, and (e)(f) are the hardness w.r.t. AdaBoost classifier.}
    \label{fig:growimb}
\end{figure*}

{\em Algorithm-level Methods}: This group of methods concentrates on modifying existing learners to alleviate their bias towards majority groups. 
It requires good insight into the modified learning algorithm and precise identification of reasons for its failure in mining skewed distributions.
The most popular algorithm-level method is cost-sensitive learning \cite{elkan2001cost-sensitive,liu2006cost-sensitive-imbalance}. By assigning large cost to minority instances and small cost to majority instances, it boosts minority importance during the learning process to alleviate the bias towards majority class.

It must be noted that the cost matrix on a specific task is given by domain expert before-hand, which is usually unavailable in many real-world problems.
Even if one has the domain knowledge required for setting the cost, such cost matrix is usally designed for specific tasks and do not generalize across different classification tasks. 
On the other hand, for the batch training models such as neural networks, the positive (minority) samples are only contained in a few batches. Even if we apply cost-sensitive into the training process, the model still soon stuck into local minima.

{\em Ensemble Methods:} This group of methods concentrates on merging one of the data-level or algorithm-level solutions with an ensemble learning method to get a robust and strong classifier. 
Ensemble approaches are gaining more popularity in real-world applications for their good performance on imbalanced tasks. 
Most of them are based on a canonical ensemble learning algorithm with an imbalance learning algorithm embedded in the pipeline, e.g., SMOTE~\cite{chawla2002smote} + Adaptive Boosting~\cite{freund1997adaboost} = SMOTEBoost~\cite{chawla2003smoteboost}.
Some other ensemble methods introduce another ensemble classifier as their base learner, e.g., EasyEnsemble~\cite{liu2009ee-bc} trains multiple AdaBoost classifier to form its final ensemble.

However, those ensemble-based methods suffer from low efficiency, poor applicability and high sensitivity to noise when applying on realistic imbalanced tasks, since they still have those data-level/algorithm-level imbalance learning methods in their pipeline.
There are few methods carried out preliminary exploration of using training feedback information to perform dynamic re-sampling on imbalance datasets.
However, such methods do not take full account of the data distribution.
For instance, BalanceCascade iteratively discards majority samples that were well-classified by the current classifier. It may result in overweighting outliers in late iterations and finally deteriorate the ensemble. 
\section{Classification Hardness Distribution}

Before we describe our algorithm, we introduce the concept of the ``classification hardness" in this section. We explain the benefits of considering hardness distribution into imbalance learning framework. We also present an intuitive visualization in Fig. \ref{fig:growimb} to help understand the relationship between hardness, imbalance ratio, class overlapping and model capacity.

{\em Definition:}
We use the symbol $\mathcal{H}$ to denote the classification hardness function, where $\mathcal{H}$ can be any ``decomposable'' error function, i.e., the overall error is calculated by the summation of individual sample errors. Examples include Absolute Error, Squared Error (Brier-score) and Cross Entropy.
Suppose $F$ is a trained classifier, we use $F(x)$ to denote the classifier's output probability of $x$ being a positive instance.
Then the classification hardness of sample $(x, y)$ with respect to $F$ is given by the function $\mathcal{H}(x, y, F)$.

{\em Advantages:} The concept of the classification hardness has two advantages under the imbalance classification scenario:

\begin{itemize}[leftmargin=0.14in]
    \item First, it fills the gap between the imbalance ratio and the task difficulty. As mentioned in the introduction, even with the same imbalance ratio, different tasks could demonstrate extremely different difficulties. We show a detailed example in Fig. \ref{fig:growimb}. In  Fig. \ref{fig:growimb_nop},  the dataset is generated with two disjoint Gaussian components. The growth of the imbalance ratio does not affect much of the task hardness. While in  Fig. \ref{fig:growimb_op} the dataset is generated by several overlapped Gaussian components. As the imbalance ratio grows, it varies from an easy classification task to an extremely hard task. However, the imbalance ratio could not well reflect such task hardness. Instead, we show the classification hardness of those two datasets based on different classifiers. As the imbalance ratio grows, the quantity of the hard samples increases sharply in Fig. \ref{fig:growimb_op_knn} and Fig. \ref{fig:growimb_op_ada}, while stays constant in  Fig. \ref{fig:growimb_nop_knn} and Fig. \ref{fig:growimb_nop_ada}. Thus, the classification hardness carries more information about the underlying data structure and better indicates the current task hardness.
    \item Second, the classification hardness also fills the gap between data sampling strategy and the classifiers' capacity. Most of the existing sampling method totally ignores the capacity of the base classifier. However, different classifiers usually demonstrate very different performances on the imbalanced data classification. For example, in Fig.\ref{fig:growimb}, KNN and Adaboost show very different hardness distribution for the same dataset. It is beneficial to consider the model capacity when performing under-sampling. Using the classification hardness, our framework is able to optimize any kind of classifier's final performance in a model-specific way.
\end{itemize}

{\em Types of Data Samples:} Intuitively, we distinguish three kinds of data samples, i.e., {\em trivial}, {\em noise} and {\em borderline} samples according to their corresponding hardness values:

\begin{itemize}[leftmargin=0.14in]
    \item Most of the data samples are {\em trivial} samples and can be well-classified by the current model, e.g., the blue samples in Fig. \ref{fig:growimb_op_knn} and Fig. \ref{fig:growimb_op_ada}. Each of the trivial samples only contributes tiny hardness. However, the overall contribution is non-negligible due to its large population. For such kind of samples, we only need to keep a small proportion of them to represent the ``skeleton" of their corresponding distribution in order to prevent overfitting, then drop most of them since they have already been well-learned.
    \item On the contrary, there are also several {\em noise} samples, e.g., the dark red samples in Fig. \ref{fig:growimb}. Despite their small population, each of them contributes a large hardness value. Thus, the total contribution can be very huge. We stress that these noise samples are usually caused by the indistinguishable overlapping or outliers since they exist stably even when the model is converged. Enforcing model to learn such samples could lead to serious overfitting. 
    \item For the rest samples, here we simply classify them as the {\em borderline} samples. The borderline samples are the most informative data samples during the training. For example, as we can see, in Fig. \ref{fig:growimb}, the light red points are very close to the decision boundary of the current model. Enlarging the weights of those borderline samples is usually helpful to further improve the model performance.
\end{itemize}

The above discussion provides us with an intuition to distinguish different data samples. However, since it is hard to make such an explicit distinction in practice,  we alternatively categorize the data samples in a ``soft'' way, as described in the next section.
\begin{figure*}[t]
    \centering
    \subfigure[Original majority set $\mathcal{N}$]{
        \label{fig:hdns_origin}
        \includegraphics[width=0.22\linewidth]{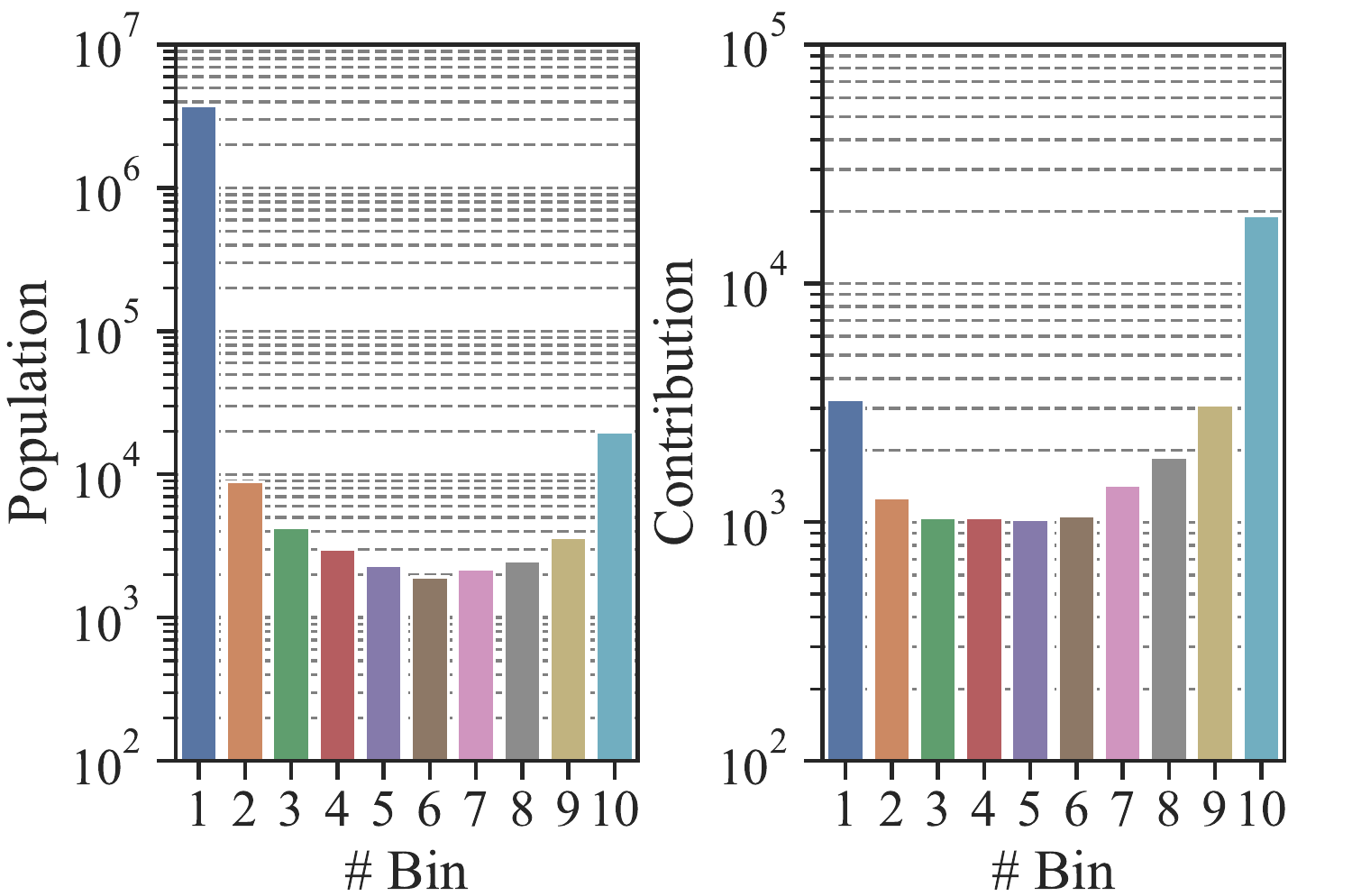}
    }
    \subfigure[$\alpha = 0$]{
        \label{fig:hdns_0}
        \includegraphics[width=0.22\linewidth]{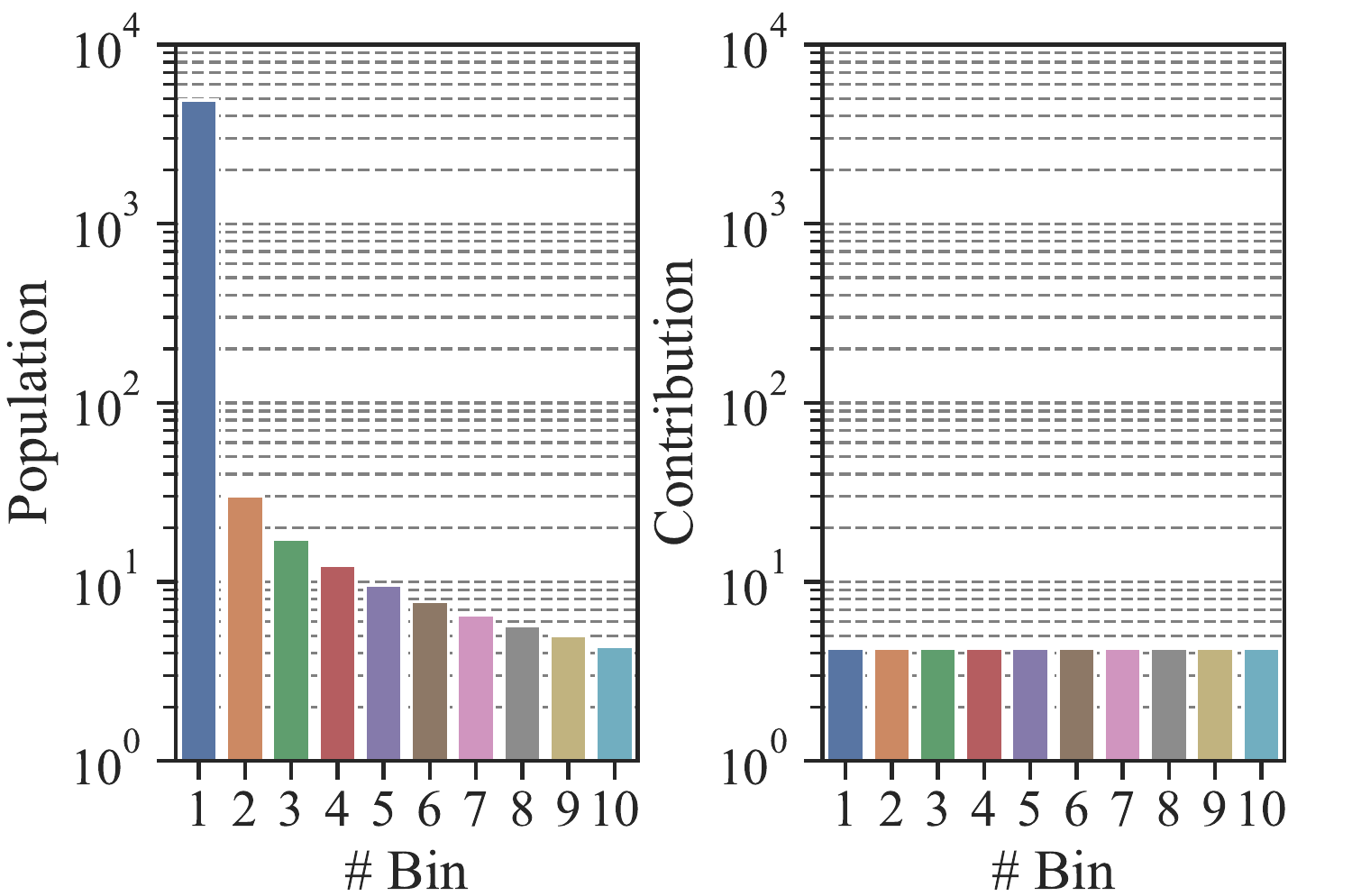}
    }
    \subfigure[$\alpha = 0.1$]{
        \label{fig:hdns_01}
        \includegraphics[width=0.22\linewidth]{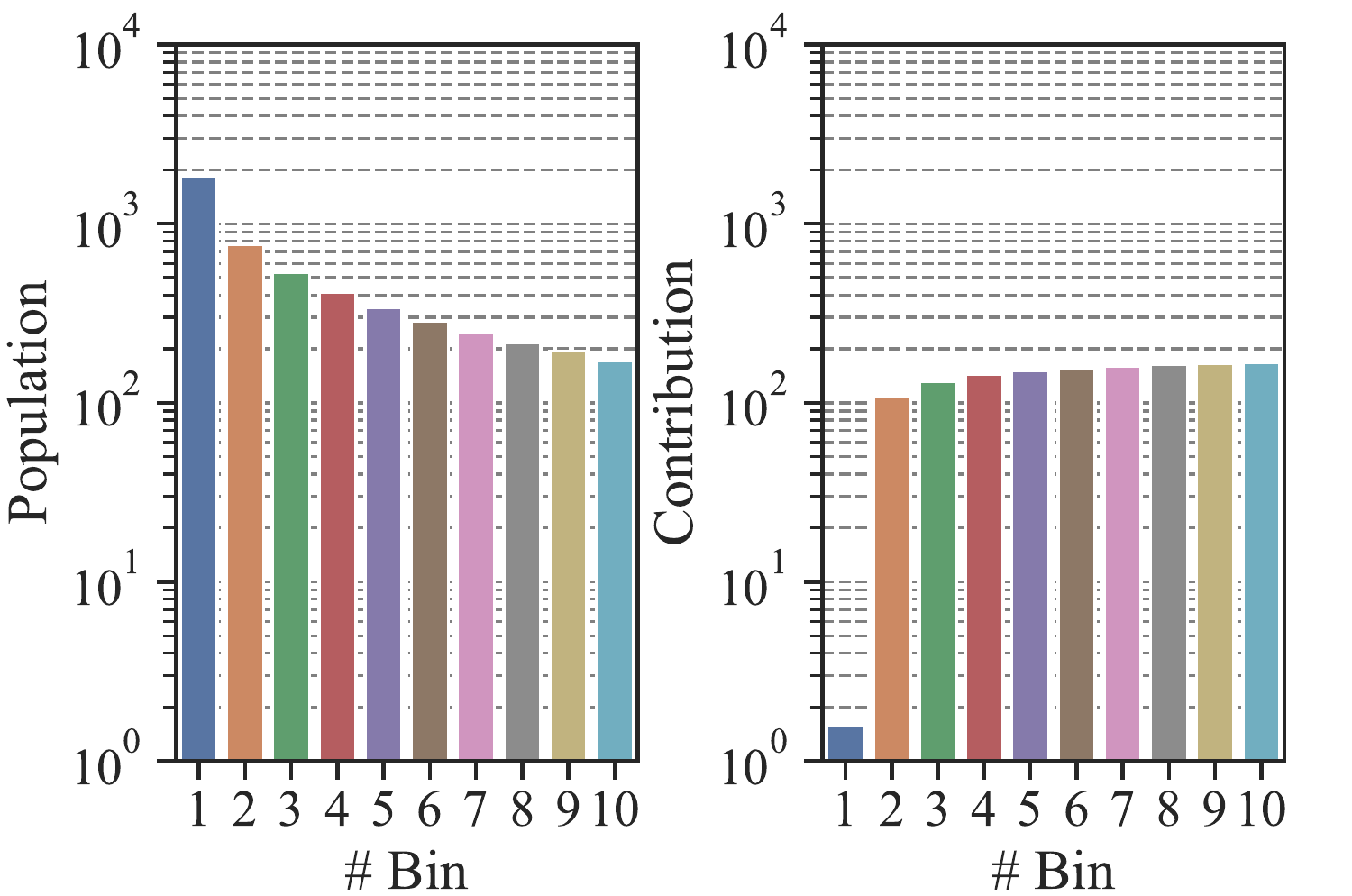}
    }
    \subfigure[$\alpha \to \infty$]{
        \label{fig:hdns_inf}
        \includegraphics[width=0.22\linewidth]{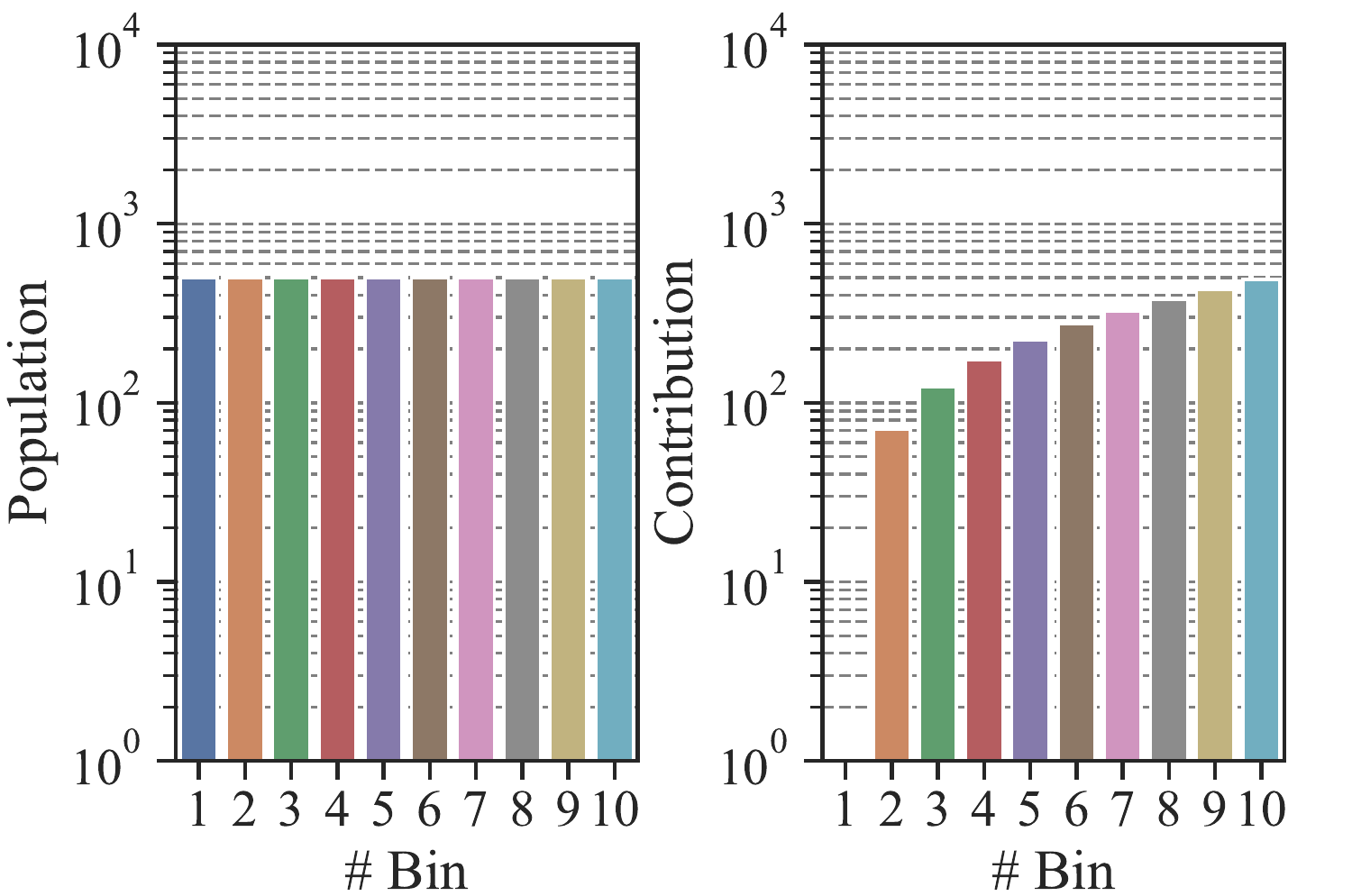}
    }
    \caption{
    An example to visualize how self-paced factor $\alpha$ controls the self-paced under-sampling. The left part of each subfigure shows the number of samples in each bin, the right part shows the hardness contribution from each bin. Subfigure (a) is the distribution over all majority instances. (b)(c)(d) are the distribution over subsets under-sampled by our mechanism when $\alpha=0$, $\alpha=0.1$, and $\alpha\to\infty$, respectively. Note that the y-axis uses log scale since the number of samples within different hardness bins can differ by orders of magnitude.
    }
    \label{fig:hdns_alpha}
\end{figure*}

\section{Self-paced Ensemble}

We now describe {\em Self-paced Ensemble}\footnote{
    \texttt{github.com/ZhiningLiu1998/self-paced-ensemble}
} (\texttt{SPE}), our framework for massive imbalance classification. Firstly, we demonstrate the ideas of hardness harmonize and self-paced factor.
After that, we summarize the \texttt{SPE} procedure in Algorithm \ref{alg:self-paced-ensemble}.

\subsection{Self-paced Under-sampling}

Motivated by previous observations, we aim to design an under-sampling mechanism that reduces the effect of trivial and noise samples, while enlarges the importance of the borderline samples as we expected. Therefore, we introduce the concept of ``hardness harmonize'' and a self-paced training procedure, to achieve such goal.

\subsubsection{Hardness Harmonize}
We split all the majority samples into $k$ bins according to their hardness values, where $k$ is a hyper-parameter. Each bin indicates a specific hardness level. 
We then under-sample the majority instances into a balanced dataset by keeping the total hardness contribution in each bin as the same. Such method is so-called ``harmonize" in the gradient-based optimization literature \cite{li2018ghm}, where they harmonize the gradient contribution in batch training of neural networks. In our case, we adopt a similar idea to harmonize the hardness in the first iteration.

However, we do not simply use the hardness harmonize in all the iterations. The main reason is that the population of trivial samples grows during the training process since the ensemble classifier will gradually fit the training set.
Hence, simply harmonizing the hardness contribution still leaves a lot of trivial samples (Fig. \ref{fig:hdns_0}). Those samples greatly slow down the learning procedure in the later iterations since they are less informative. Instead, we introduce the ``self-pace factor'' to perform self-paced harmonize under-sampling.

\subsubsection{Self-paced Factor}
Specifically, start from harmonizing the hardness contribution of each bin, we gradually decrease the sample probability of those bins with a large population.
The decreasing level is controlled by a self-paced factor $\alpha$.
When $\alpha$ goes large, we focus more on the harder samples instead of the simple hardness contribution harmonize.
In the first few iterations, our framework mainly focuses on those informative borderline samples, thus the outliers and noises do not affect much of the generalization ability of our model. In the later iterations where $\alpha$ is very large, our framework still keeps a reasonable fraction of trivial (high confidence) samples as the ``skeleton", which effectively prevents our framework from overfitting.
Fig. \ref{fig:hdns_alpha} shows the self-paced under-sampling process of a real-world large-scale dataset\footnote{Payment Simulation dataset, statistics can be found in Table \ref{datasets}.
}.

\subsection{Algorithm Formalization}
Finally, in this subsection, we describe our algorithm formally.
Recall that in Section 2, we use $\mathcal{D}$ to denote the collection of all training samples $(x, y)$. 
$\mathcal{N}$ / $\mathcal{P}$ is the majority / minority set in $\mathcal{D}$.
We use $\mathcal{D}_{dev}$ to denote the validation set, which is used to measure the generalization ability of the ensemble model. Note that $\mathcal{D}_{dev}$ keeps the original imbalanced distribution with no re-sampling.
Moreover, we use $B_\ell$ to denote the $\ell$-th bin, where $B_\ell$ is defined as 
$$B_\ell = \{(x, y) \ |\ \frac{\ell - 1}{k} \leq \mathcal{H}(x,y,F) < \frac{\ell}{k}\} \ w.l.o.g.\ \mathcal{H}\in[0, 1]$$
The details are shown in Algorithm \ref{alg:self-paced-ensemble}. 
Notice that we update hardness value in each iteration (line 4-5) in order to select data samples that were most beneficial for the current ensemble.
We use the $\mathrm{tan}$ function (line 7) to control the growth of self-paced factor $\alpha$. 
Thus we have $\alpha=0$ in the first iteration and $\alpha\to\infty$ in the last iteration.

\begin{algorithm}[h]
\caption{Self-paced Ensemble}
\label{alg:self-paced-ensemble}
\LinesNumbered 
\KwIn{
    Training set $\mathcal{D}$,
    hardness function $\mathcal{H}$,
    base classifier $f$,
    number of base classifiers $n$,
    number of bins $k$,
}
\textbf{Initialize:} 
    $\mathcal{P} \Leftarrow $ minority in $\mathcal{D}$, 
    $\mathcal{N} \Leftarrow $ majority in $\mathcal{D}$

Train classifier $f_{0}$ using random under-sample majority subsets $\mathcal{N}'_{0}$ and $\mathcal{P}$, where $|\mathcal{N}'_{0}|=|\mathcal{P}|$.

\For{i=1 to $n$}{
    
    Ensemble $F_{i}(x) = \frac{1}{i} \sum\limits_{j=0}^{i-1} f_{j}(x)$
    
    Cut majority set into $k$ bins w.r.t. $\mathcal{H}(x, y, F_{i})$:
    $B_{1}, B_{2}, \cdots , B_{k}$
    
    Average hardness contribution in $\ell$-th bin:
    $h_\ell = \sum_{s \in B_{\ell}} \mathcal{H}(x_s, y_s, F_{i}) / |B_\ell|, \ \forall \ell = 1, \ldots, k$
    
    Update self-paced factor $\alpha = tan(\frac{i\pi}{2n})$

    Unnormalized sampling weight of $\ell$-th bin: $p_\ell = \frac{1}{h_\ell + \alpha}, \forall \ \ell = 1, \ldots, k$
    
    Under-sample from $\ell$-th bin with $\frac{p_\ell}{\sum_m {p_m}} \cdot |\mathcal{P}|$ samples 
    
    Train $f_{i}$ using newly under-sampled subset 
    
}

\Return{ final ensemble $F(x) = \frac{1}{n}\sum_{m=1}^n f_m(x)$}
\end{algorithm}

\begin{table*}[hbt!]
\scriptsize
\renewcommand{\arraystretch}{1}
\caption{
Generalized performance (AUCPRC) on checker board dataset.
}
\label{result_checkerboard}
\centering

\begin{tabular}{|bc|tc|tctctctctc|tc|}
\hline
\rowstyle{\bfseries}
Model & Hyper & \texttt{RandUnder} & \texttt{Clean} & \texttt{SMOTE} & $\texttt{Easy}_{10}$ & $\texttt{Cascade}_{10}$ & $\texttt{SPE}_{10}$ \\
\hline
\multirow{1}*{KNN} & \multirow{1}*{k\_neighbors=5}
 & 0.281$\pm$0.003 & 0.382$\pm$0.000 & 0.271$\pm$0.003 & 0.411$\pm$0.003 & 0.409$\pm$0.005 & {\bf 0.498}$\pm$0.004\\
\multirow{1}*{DT} & \multirow{1}*{max\_depth=10}
 & 0.236$\pm$0.010 & 0.365$\pm$0.001 & 0.299$\pm$0.007 & 0.463$\pm$0.009 & 0.376$\pm$0.052 & {\bf 0.566}$\pm$0.011\\
\multirow{1}*{MLP} & \multirow{1}*{hidden\_unit=128}
 & 0.562$\pm$0.017 & 0.138$\pm$0.035 & 0.615$\pm$0.009 & 0.610$\pm$0.004 & 0.582$\pm$0.005 & {\bf 0.656}$\pm$0.005\\
\multirow{1}*{SVM} & \multirow{1}*{C=1000}
 & 0.306$\pm$0.003 & 0.405$\pm$0.000 & 0.324$\pm$0.002 & 0.386$\pm$0.001 & 0.456$\pm$0.010 & {\bf 0.518}$\pm$0.004\\
\multirow{1}*{AdaBoost$_{10}$} & \multirow{1}*{n\_estimator=10}
 & 0.226$\pm$0.019 & 0.362$\pm$0.000 & 0.297$\pm$0.004 & 0.487$\pm$0.017 & 0.391$\pm$0.013 & {\bf 0.570}$\pm$0.008\\
\multirow{1}*{Bagging$_{10}$} & \multirow{1}*{n\_estimator=10}
 & 0.273$\pm$0.002 & 0.401$\pm$0.000 & 0.316$\pm$0.003 & 0.436$\pm$0.004 & 0.389$\pm$0.007 & {\bf 0.568}$\pm$0.005\\
\multirow{1}*{RandForest$_{10}$} & \multirow{1}*{n\_estimator=10}
 & 0.260$\pm$0.004 & 0.229$\pm$0.000 & 0.306$\pm$0.011 & 0.454$\pm$0.005 & 0.402$\pm$0.012 & {\bf 0.572}$\pm$0.003\\
\multirow{1}*{GBDT$_{10}$} & \multirow{1}*{boost\_rounds=10}
 & 0.553$\pm$0.015 & 0.602$\pm$0.000 & 0.591$\pm$0.008 & 0.645$\pm$0.006 & 0.648$\pm$0.009 & {\bf 0.680}$\pm$0.003\\
\hline
\end{tabular}
\end{table*}

\section{Experiments \& Analysis}

In this section, we present the results of our experimental study on one synthetic and five real-world extremely imbalanced datasets. We tested the applicability of our proposed algorithm to incorporate with different kinds of base classifiers. We also show some visualizations to help understand the difference between our proposed method and the other imbalance learning methods. We evaluated the experiment results with multiple criteria, and demonstrate the strength of our proposed framework. 

\subsection{Synthetic Dataset}
\begin{figure}[h]
    \centering
    \includegraphics[width=0.9\linewidth]{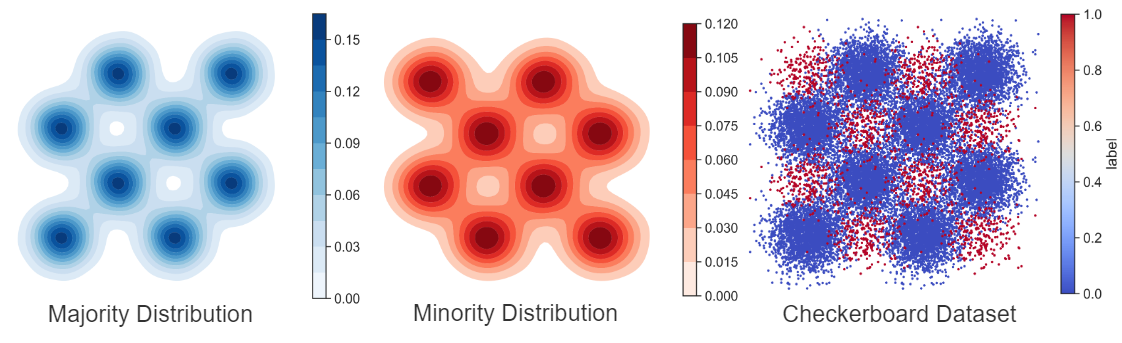}
    \caption{An example of {\em checkerboard} dataset. Blue dots represent the majority class samples, red ones represent the minority class samples.}
    \label{fig:checker_board}
\end{figure}

To provide more insights of our framework, we first show the experimental results on the synthetic dataset. We create a $4 \times 4$ {\em checkerboard} dataset to validate our method. The dataset contains $16$ Gaussian components. All Gaussian components share the same covariance matrix of $0.1 \times \mathbf{I}_2$. We set the number of minority samples $|\mathcal{P}|$ as $1,000$, and the number of majority $|\mathcal{N}|$ as $10,000$. The training set $\mathcal{D}$, validation set $\mathcal{D}_{dev}$ and test set $\mathcal{D}_{test}$ were independently sampled from same original distribution. See Fig. \ref{fig:checker_board} for an example.

\subsubsection{Setup Details}
We compared our proposed method \texttt{SPE}\footnote{In our implementation of \texttt{SPE}, we set the number of bins $k=20$, and use absolute error as the classification hardness, i.e., $\mathcal{H}(x,y,F)=|F(x)-y|$, unless otherwise stated.} with following imbalance learning approaches:

\begin{itemize}[leftmargin=0.14in, label=-]
    \item \texttt{RandUnder} ({\em Random Under-sampling}) randomly under-sample the majority class to get a subset $\mathcal{N}'$ such that $|\mathcal{N}'|=|\mathcal{P}|$. The set $\mathcal{N}'\cup\mathcal{P}$ was then used for training.
    \item \texttt{Clean} ({\em Neighbourhood Cleaning Rule based under-sampling})~\cite{laurikkala2001ncr} removes a majority instance if most of its neighbors come from another class.
    \item \texttt{SMOTE} ({\em Synthetic Minority Over-sampling TechniquE})~\cite{chawla2002smote} generates synthetic minority instances between existing minority samples until the dataset is balanced.
    \item \texttt{Easy} ({\em EasyEnsemble})~\cite{liu2009ee-bc} utilizes \texttt{RandUnder} to train multiple AdaBoost~\cite{freund1997adaboost} models and combine their outputs.
    \item \texttt{Cascade} ({\em BalanceCascade})~\cite{liu2009ee-bc} extends \texttt{Easy} by iteratively drop majority examples that were already well classified by current base classifier.
\end{itemize}

In addition, according to our aforementioned discussion in the Classification Hardness section, by considering the hardness distribution our proposed framework \texttt{SPE} is able to work with any kind of classifiers and optimize the final performance in a model-specific way.
Hence, we introduce 8 canonical classifiers in order to test the effectiveness and applicability of different imbalance learning methods:

\begin{itemize}[label=-]
    \item {\em K-Nearest Neighbors} (KNN) \cite{altman1992knn}
    \item {\em Decision Tree} (DT) \cite{quinlan1986dt}
    \item {\em Support Vector Machine} (SVM) \cite{cortes1995svm}
    \item {\em Multi-Layer Perceptron} (MLP) \cite{haykin2009neuralnetworks}
    \item {\em Adaptive Boosting} (AdaBoost) \cite{freund1997adaboost}
    \item {\em Bootstrap aggregating} (Bagging) \cite{breiman1996bagging}
    \item {\em Random Forest} (RandForest) \cite{liaw2002rf}
    \item {\em Gradient Boosting Decision Tree} (GBDT) \cite{friedman2002gbdt}
\end{itemize}

We apply imbalanced-learn~\cite{guillaume2017imblearn} package to implement aforementioned imbalance learning methods, and scikit-learn~\cite{pedregosa2011sklearn}, LightGBM~\cite{ke2017lightgbm}, Pytorch~\cite{paszke2017pytorch} packages to implement the canonical classifiers. 
We use subscripts to denote the number of base models in an ensemble classifier, e.g., \texttt{Easy}$_{10}$ indicates \texttt{Easy} with 10 base models.
Due to space limitation, we only present the experimental results of AUCPRC in this experiment, other metrics will be used in following extensive experiments on real-world datasets.

\subsubsection{Results on synthetic dataset}

Table \ref{result_checkerboard} lists the results on checkerboard task. 
Note that to reduce randomness, we show the mean and standard deviation of 10 independent runs.
We also list the hyper-parameters we used for each base classifier.
From the Table \ref{result_checkerboard} we can observe that:
\begin{itemize}[leftmargin=0.14in]
    \setlength{\itemsep}{0pt}
    \setlength{\parsep}{0pt}
    \item \texttt{SPE} consistently outperform other methods on the checkerboard dataset using 8 different classifiers.
    \item 
    Distance-based re-sampling lead to poor results when cooperating with specific classifiers, e.g., \texttt{SMOTE}+KNN, \texttt{Clean}+RandForest. 
    We argue that the ignorance of difference in model capacity is the main reason that causes invalidity to those re-sample methods. 
    \item Comparing with other methods, ensemble methods \texttt{Easy} and \texttt{Cascade} obtain better and more robust performance but still worse than our proposed ensemble framework \texttt{SPE}.
\end{itemize}


\begin{figure}[h]
    \centering
    \subfigure[$cov = 0.05$]{
        \label{fig:tc-005}
        \includegraphics[width=0.28\linewidth]{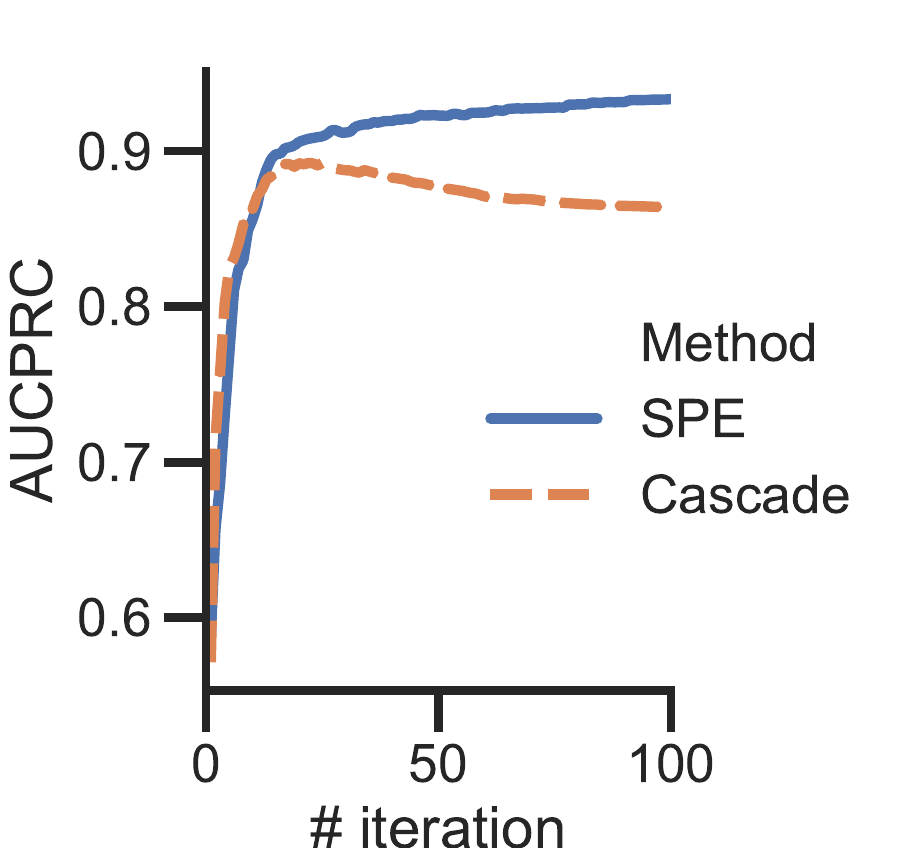}
    }
    \subfigure[$cov = 0.10$]{
        \label{fig:tc-0075}
        \includegraphics[width=0.28\linewidth]{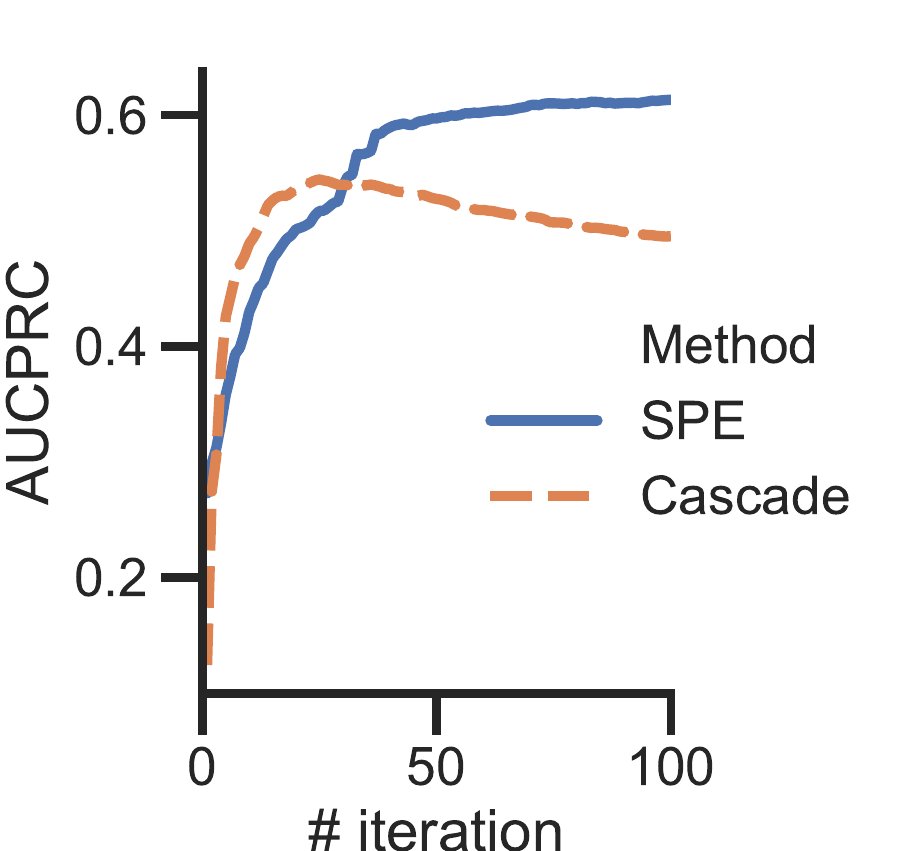}
    }
    \subfigure[$cov = 0.15$]{
        \label{fig:tc-01}
        \includegraphics[width=0.28\linewidth]{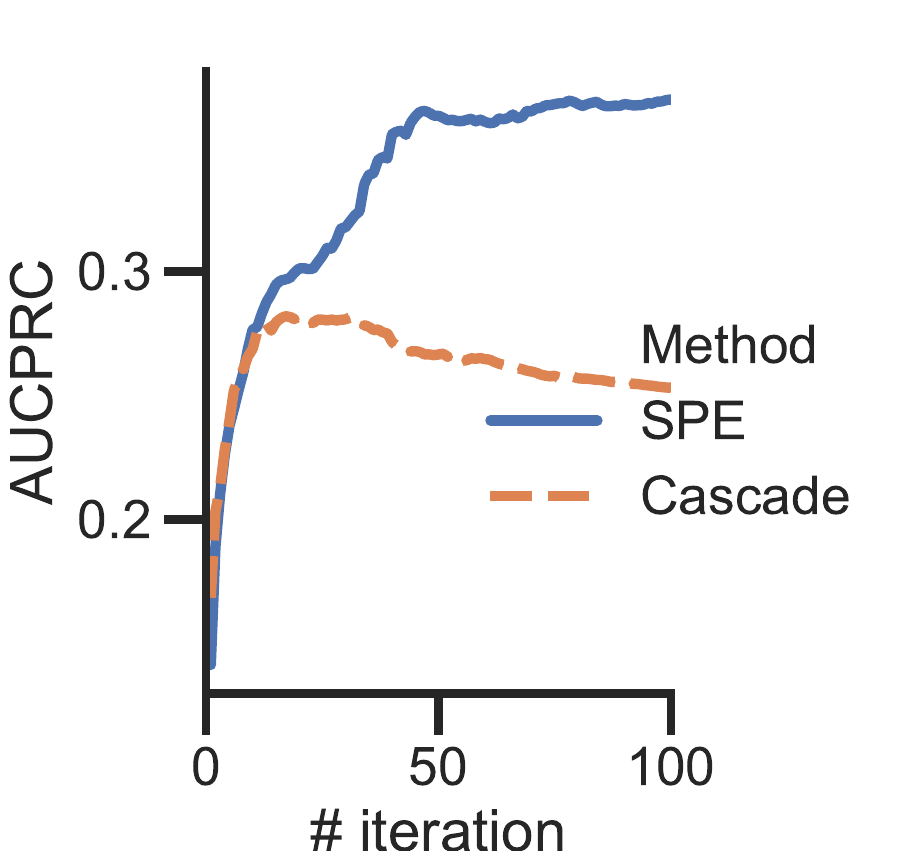}
    }
    \caption{Training curve under different level of overlap.}
    \label{fig:training-curve}
\end{figure}

\subsubsection{Robustness under Class Overlapping} Furthermore, we test the robustness of \texttt{SPE}, when the Gaussian components have different levels of overlapping. We control the components overlapping by replacing the original covariance matrix from $0.1 \times \mathbf{I}_{2}$ to $0.05 \times \mathbf{I}_{2}$ and $0.15 \times \mathbf{I}_{2}$.
The distribution is less overlapped when the covariance factor in covariance matrix is smaller, and more overlapped when it is bigger. 
We keep the size and imbalance ratio to be the same, and sample three different checkerboard datasets respectively. Fig. \ref{fig:training-curve} shows how the AUCPRC (on test set) changes within training process:

\begin{itemize}[leftmargin=0.14in]
    \setlength{\itemsep}{0pt}
    \setlength{\parsep}{0pt}
    \item The level of distribution overlapping significantly influences the classification performance, even though the size and imbalance ratio of all datasets are totally the same.
    \item As the overlapping aggravates, the performance of \texttt{Cascade} shows more obvious downward trend in later iterations. The reason behind is that \texttt{Cascade} inclines to overfit the noise samples, while \texttt{SPE} can alleviate this issue by keeping a reasonable proportion of trivial and borderline samples.
\end{itemize}

\begin{table*}[h]
\scriptsize
\renewcommand{\arraystretch}{1.0}
\caption{Statistics of the real-world datasets}
\label{datasets}
\centering
\begin{tabular}{|bc|tctc|tctctc|}
\hline
\rowstyle{\bfseries}
Dataset         & \#Attribute  & \#Sample & Feature Format            & Imbalance Ratio     & Model\\
\hline
Credit Fraud    & 31            & 284,807   & Numerical                 & 578.88:1          & KNN, DT, MLP\\
KDDCUP (DOS vs. PRB) & 42       & 3,924,472 & Integer \& Categorical    & 94.48:1           & AdaBoost$_{10}$\\
KDDCUP (DOS vs. R2L) & 42       & 3,884,496 & Integer \& Categorical    & 3448.82:1         & AdaBoost$_{10}$\\
Record Linkage  & 12            & 5,749,132 & Numerical \& Categorical  & 273.67:1          & GBDT$_{10}$\\
Payment Simulation  & 11            & 6,362,620 & Numerical \& Categorical  & 773.70:1          & GBDT$_{10}$\\
\hline
\end{tabular}
\end{table*}

\begin{table*}[h]
\scriptsize
\renewcommand{\arraystretch}{1.0}
\caption{Generalized performance on 5 real-world datasets.}
\label{result-real}
\centering
\begin{tabular}{|bc|tc|tc|tctctctctc|tc|}
\hline
\rowstyle{\bfseries}
Dataset & Model & Metric & \texttt{RandUnder} & \texttt{Clean} & \texttt{SMOTE} & $\texttt{Easy}_{10}$ & $\texttt{Cascade}_{10}$ & $\texttt{SPE}_{10}$ \\
\hline
\multirow{12}*{Credit Fraud} 
 & \multirow{4}*{KNN} 
   & AUCPRC	& 0.052$\pm$0.002 & 0.677$\pm$0.000 & 0.352$\pm$0.000 & 0.162$\pm$0.012 & 0.676$\pm$0.015 & {\bf 0.752}$\pm$0.018\\
 & & F1 	& 0.112$\pm$0.007 & 0.821$\pm$0.000 & 0.559$\pm$0.001 & 0.250$\pm$0.020 & 0.792$\pm$0.023 & {\bf 0.843}$\pm$0.016\\
 & & GM 	& 0.228$\pm$0.009 & 0.822$\pm$0.000 & 0.593$\pm$0.001 & 0.399$\pm$0.025 & 0.810$\pm$0.001 & {\bf 0.852}$\pm$0.002\\
 & & MCC 	& 0.222$\pm$0.014 & 0.822$\pm$0.000 & 0.592$\pm$0.001 & 0.650$\pm$0.004 & 0.815$\pm$0.006 & {\bf 0.855}$\pm$0.006\\
 \cline{2-9}
 & \multirow{4}*{DT}
   & AUCPRC & 0.014$\pm$0.001 & 0.598$\pm$0.013 & 0.088$\pm$0.011 & 0.339$\pm$0.039 & 0.592$\pm$0.029 & {\bf 0.783}$\pm$0.015\\
 & & F1 	& 0.032$\pm$0.002 & 0.767$\pm$0.004 & 0.177$\pm$0.006 & 0.478$\pm$0.021 & 0.737$\pm$0.023 & {\bf 0.838}$\pm$0.021\\
 & & GM 	& 0.119$\pm$0.003 & 0.778$\pm$0.006 & 0.303$\pm$0.017 & 0.548$\pm$0.048 & 0.749$\pm$0.011 & {\bf 0.843}$\pm$0.007\\
 & & MCC 	& 0.124$\pm$0.001 & 0.780$\pm$0.008 & 0.310$\pm$0.003 & 0.409$\pm$0.015 & 0.778$\pm$0.049 & {\bf 0.831}$\pm$0.008\\
 \cline{2-9}
 & \multirow{4}*{MLP}
   & AUCPRC & 0.225$\pm$0.050 & 0.001$\pm$0.000 & 0.527$\pm$0.017 & 0.605$\pm$0.016 & 0.738$\pm$0.009 & {\bf 0.747}$\pm$0.011\\
 & & F1 	& 0.388$\pm$0.047 & 0.003$\pm$0.000 & 0.725$\pm$0.013 & 0.762$\pm$0.023 & 0.803$\pm$0.004 & {\bf 0.811}$\pm$0.010\\
 & & GM 	& 0.494$\pm$0.040 & 0.040$\pm$0.000 & 0.665$\pm$0.060 & 0.748$\pm$0.019 & 0.806$\pm$0.007 & {\bf 0.828}$\pm$0.003\\
 & & MCC 	& 0.178$\pm$0.008 & 0.000$\pm$0.000 & 0.718$\pm$0.006 & 0.705$\pm$0.004 & 0.744$\pm$0.046 & {\bf 0.826}$\pm$0.008\\
\hline

                & \multirow{4}*{AdaBoost$_{10}$} 
                & AUCPRC    & 0.930$\pm$0.012 & - - - & - - - & 0.995$\pm$0.002 & {\bf 1.000}$\pm$0.000 & {\bf 1.000}$\pm$0.000\\
KDDCUP          & & F1 	    & 0.962$\pm$0.001 & - - - & - - - & 0.997$\pm$0.000 & {\bf 0.999}$\pm$0.000 & {\bf 0.999}$\pm$0.000\\
(DOS vs. PRB)   & & GM 	    & 0.964$\pm$0.001 & - - - & - - - & 0.997$\pm$0.001 & 0.998$\pm$0.000 & {\bf 0.999}$\pm$0.000\\
                & & MCC     & 0.956$\pm$0.004 & - - - & - - - & 0.992$\pm$0.001 & 0.993$\pm$0.003 & {\bf 0.999}$\pm$0.000\\
\hline

                & \multirow{4}*{AdaBoost$_{10}$} 
                & AUCPRC    & 0.034$\pm$0.005 & - - - & - - - & 0.108$\pm$0.011 & 0.945$\pm$0.005 & {\bf 0.999}$\pm$0.001\\
KDDCUP          & & F1 	    & 0.050$\pm$0.005 & - - - & - - - & 0.259$\pm$0.058 & 0.965$\pm$0.005 & {\bf 0.991}$\pm$0.003\\
(DOS vs. R2L)   & & GM 	    & 0.164$\pm$0.011 & - - - & - - - & 0.329$\pm$0.015 & 0.967$\pm$0.008 & {\bf 0.988}$\pm$0.004\\
                & & MCC 	& 0.175$\pm$0.016 & - - - & - - - & 0.214$\pm$0.004 & 0.905$\pm$0.056 & {\bf 0.986}$\pm$0.004\\
\hline

\multirow{4}*{Record Linkage}
 & \multirow{4}*{GBDT$_{10}$} 
   & AUCPRC 	& 0.988$\pm$0.011 & - - - & - - - & 0.999$\pm$0.000 & {\bf 1.000}$\pm$0.000 & {\bf 1.000}$\pm$0.000\\
 & & F1 		& 0.995$\pm$0.000 & - - - & - - - & 0.996$\pm$0.000 & {\bf 0.998}$\pm$0.000 & {\bf 0.998}$\pm$0.000\\
 & & GM 		& 0.994$\pm$0.002 & - - - & - - - & 0.996$\pm$0.000 & {\bf 0.998}$\pm$0.000 & {\bf 0.998}$\pm$0.000\\
 & & MCC 		& 0.780$\pm$0.000 & - - - & - - - & 0.884$\pm$0.000 & 0.940$\pm$0.000 & {\bf 0.998}$\pm$0.000\\
\hline

\multirow{4}*{Payment Simulation}
 & \multirow{4}*{GBDT$_{10}$} 
   & AUCPRC	    & 0.278$\pm$0.030 & - - - & - - - & 0.676$\pm$0.058 & 0.776$\pm$0.004 & {\bf 0.944}$\pm$0.001\\
 & & F1 	    & 0.446$\pm$0.030 & - - - & - - - & 0.709$\pm$0.021 & 0.851$\pm$0.003 & {\bf 0.885}$\pm$0.001\\
 & & GM 	    & 0.530$\pm$0.020 & - - - & - - - & 0.735$\pm$0.011 & 0.851$\pm$0.001 & {\bf 0.885}$\pm$0.001\\
 & & MCC 	    & 0.290$\pm$0.023 & - - - & - - - & 0.722$\pm$0.015 & 0.856$\pm$0.002 & {\bf 0.876}$\pm$0.001\\
\hline

\end{tabular}
\end{table*}

\begin{figure}[hbt!]
    \centering
    \subfigure[\texttt{Clean}]{
    \begin{minipage}[b]{0.16\linewidth}
        \fbox{\includegraphics[width=1\linewidth]{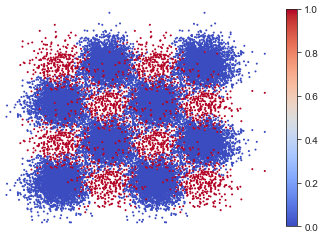}}\vspace{23pt}
        \fbox{\includegraphics[width=1\linewidth]{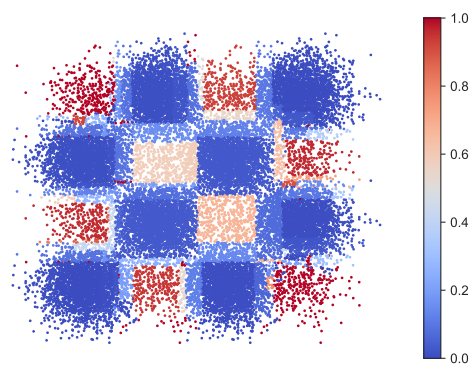}}\vspace{5pt}
    \end{minipage}
    }
    \subfigure[\texttt{SMOTE}]{
    \begin{minipage}[b]{0.16\linewidth}
        \fbox{\includegraphics[width=1\linewidth]{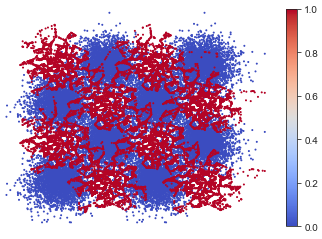}}\vspace{23pt}
        \fbox{\includegraphics[width=1\linewidth]{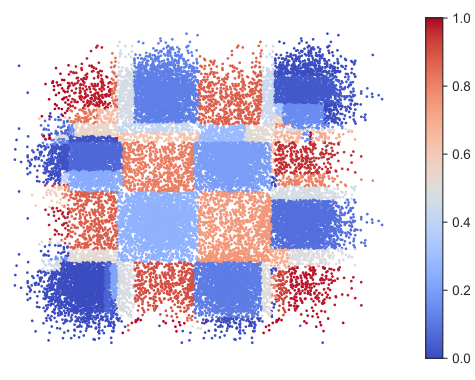}}\vspace{5pt}
    \end{minipage}
    }
    \subfigure[\texttt{Easy}]{
    \begin{minipage}[b]{0.16\linewidth}
        \fbox{\includegraphics[width=1\linewidth]{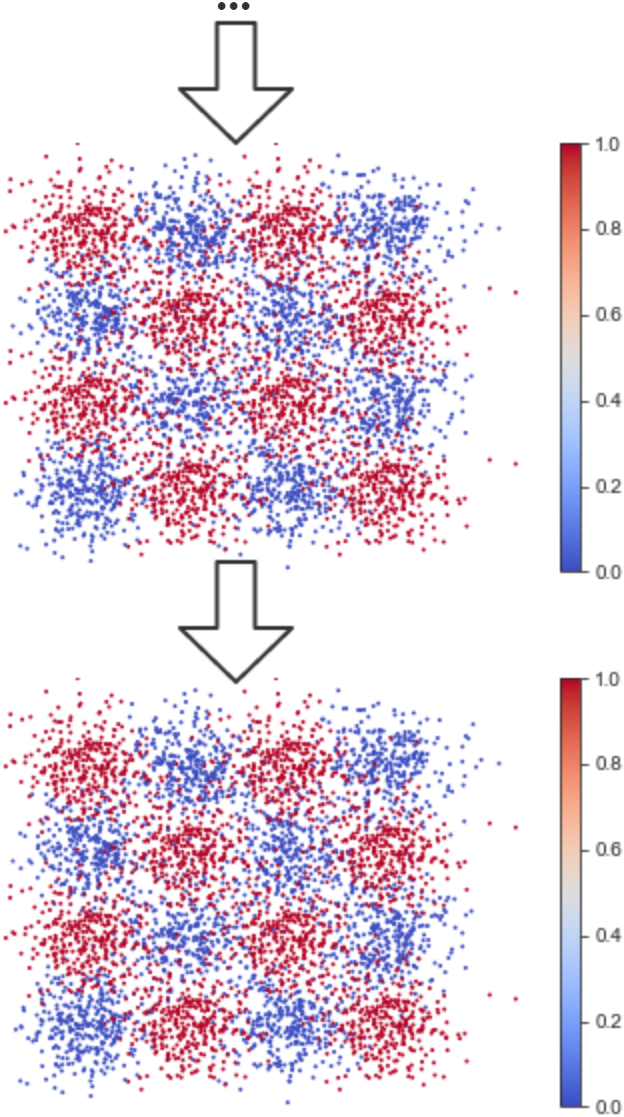}}\vspace{5pt}
        \fbox{\includegraphics[width=1\linewidth]{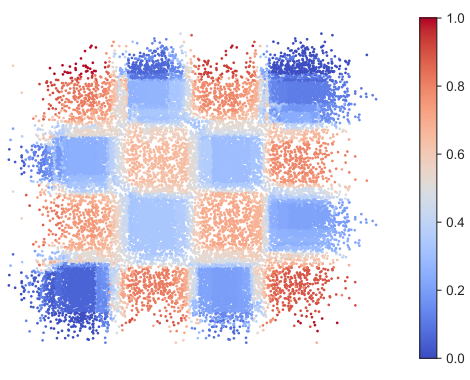}}\vspace{5pt}
    \end{minipage}
    }
    \subfigure[\texttt{Cascade}]{
    \begin{minipage}[b]{0.16\linewidth}
        \fbox{\includegraphics[width=1\linewidth]{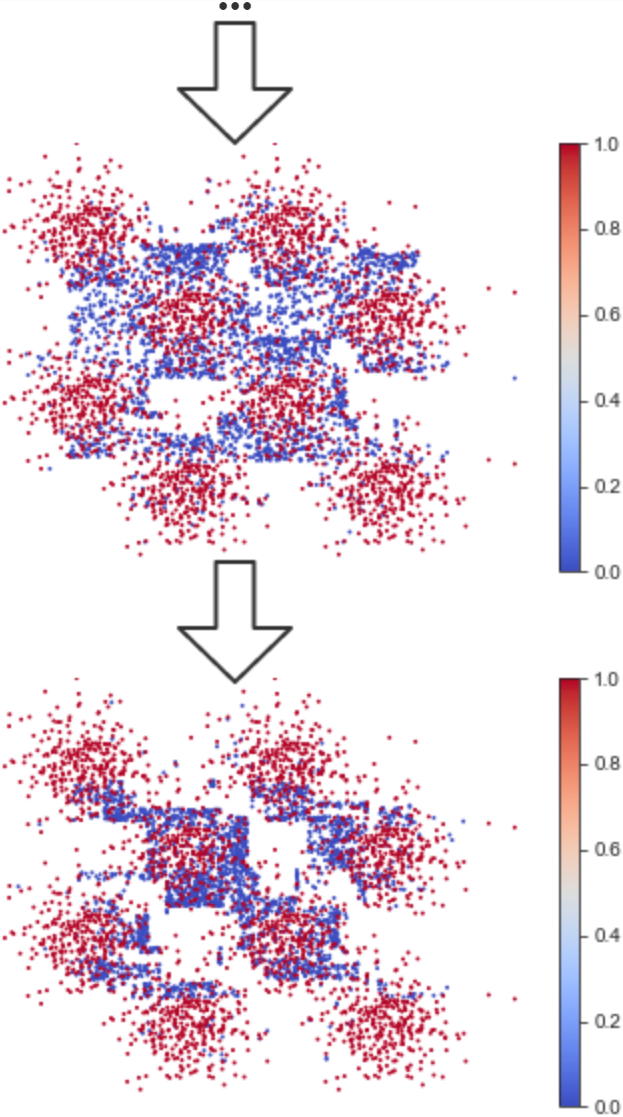}}\vspace{5pt}
        \fbox{\includegraphics[width=1\linewidth]{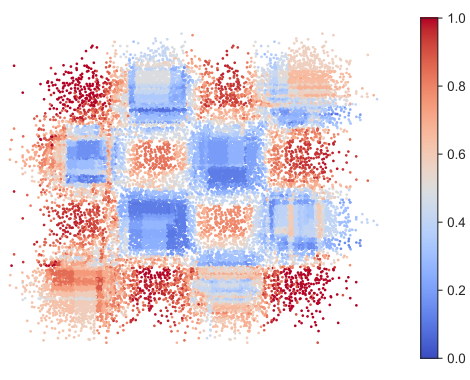}}\vspace{5pt}
    \end{minipage}
    }
    \subfigure[\texttt{SPE}]{
    \begin{minipage}[b]{0.16\linewidth}
        \fbox{\includegraphics[width=1\linewidth]{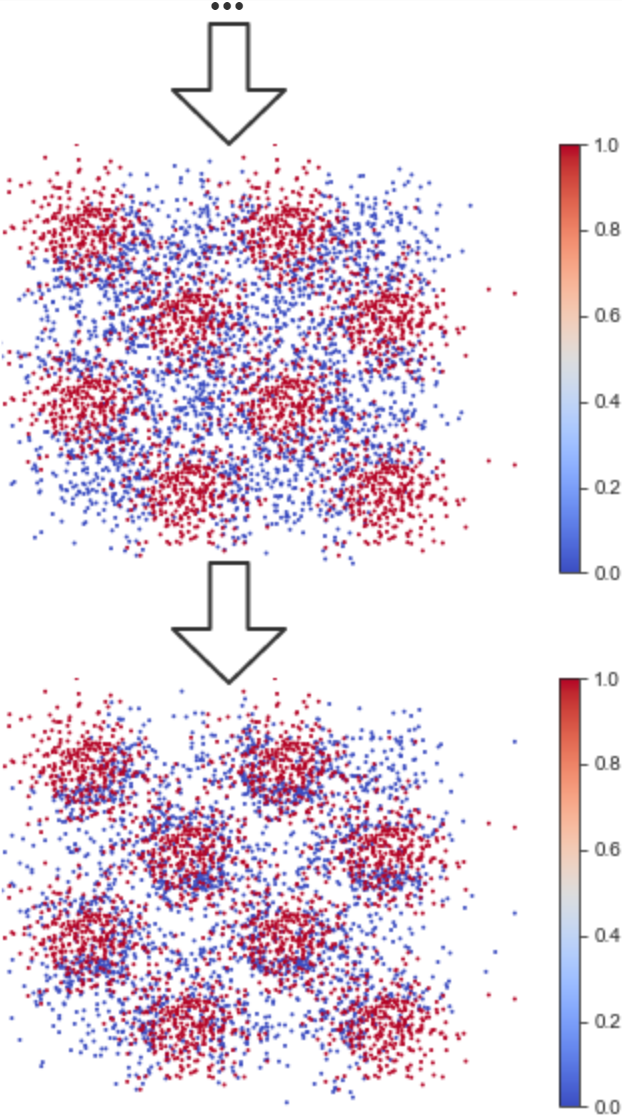}}\vspace{5pt}
        \fbox{\includegraphics[width=1\linewidth]{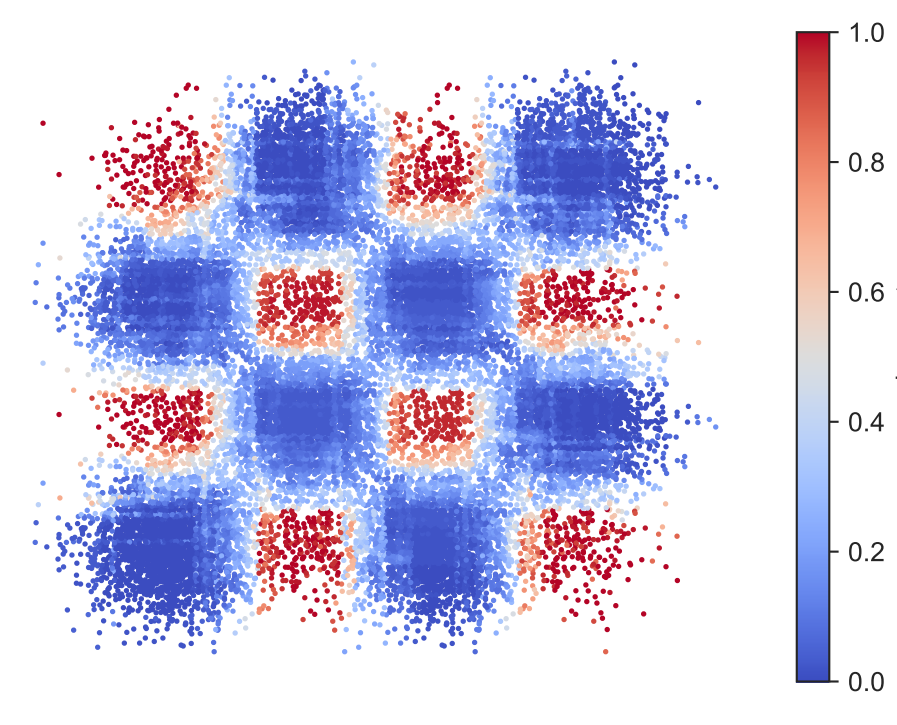}}\vspace{5pt}
    \end{minipage}
    }
    \caption{Visualization of training set (upper, blue/red dot indicates a sample from majority/minority class) and predict probability (lower, blue/red dot indicates the classifier tend to classify a sample as majority/minority class) on {\em checkerboard} dataset. Note that ensemble methods \texttt{Easy}, \texttt{Cascade} and \texttt{SPE} train multiple base models in each iteration with different training sets, so we show training sets of $5_{th}$ and $10_{th}$ model in their pipeline.}
    \label{fig:checkerboard_vis}
\end{figure}

\subsubsection{Intuitive Visualization} We give a visualization in Fig. \ref{fig:checkerboard_vis} to show how the aforementioned imbalance learning approaches train / predict on checkerboard dataset. 
As we can see, \texttt{Clean} tries to clean up the majority outliers who were surrounded by minority data points, however, it retains all the trivial samples so that the learning model cannot focus on more informative data. \texttt{SMOTE} over-generalizes minority class due to indistinguishable overlapping. \texttt{Easy} performs simple random under-sampling and thus part of majority samples are dropped which causes the information loss. \texttt{Cascade} keeps many outliers in late iterations. Those outliers finally lead to bad generalization. By contrast, \texttt{SPE} 
gets a much more accurate and robust results by considering the classification hardness distribution over the dataset.

\subsection{Real-world Datasets}
We choose several real-life datasets with highly skewed class distribution to assess the effectiveness of our proposed learning framework on realistic tasks. 

\underline{Credit Fraud} contains transactions made by credit cards in September 2013 by European card-holders \cite{dal2018creditfraud}. The task is to detect frauds from credit card transaction records. It is a highly imbalanced dataset with only 492 frauds out of 284,807 transactions, which brings a high imbalance ratio of 578.88:1.
\underline{Payment Simulation} is a large-scale dataset with 6.35 million instances. It simulates mobile money transactions based on a sample of real transactions extracted from one month of financial logs from a mobile money service implemented in an African country. Similarly, it has 8,213 frauds out of 6,362,620 transactions and a high imbalance ratio 773.70:1.
\underline{Record Linkage} is a dataset of element-wise comparison of records with personal data from a record linkage setting. The task requires us to decide whether the underlying records belong to one person. The underlying records stem from the epidemiological cancer registry of the German state of North Rhine-Westphalia, which has 5,749,132 record pairs, and 20,931 of them are matches.
\underline{KDDCUP-99} contains a standard set of data to be audited, which includes a wide variety of intrusions simulated in a military network environment. The competition task was to build a network intrusion detector, a predictive model capable of distinguishing between ``bad'' connections, called intrusions or attacks, and ``good'' normal connections. It is a multi-class task with 4 main categories of attacks: DOS, R2L, U2R and probing (PRB). We formed 2 two-class imbalanced problems by taking the majority class (i.e., DOS) and a minority class (i.e., PRB and R2L), namely, \underline{KDDCUP (DOS vs. PRB)} and \underline{KDDCUP (DOS vs. R2L)}.

Table \ref{datasets} lists the statistics of each dataset.

\begin{table*}[h]
\scriptsize
\renewcommand{\arraystretch}{1.0}
\caption{Generalized performance (AUCPRC) of 12 different re-sampling methods.}
\label{result-resample}
\centering
\begin{tabular}{|bc|tc|tctctctctc|tc|tc|}
\hline
\rowstyle{\bfseries}
Category & Method & LR & KNN & DT & AdaBoost$_{10}$  & GBDT$_{10}$ & \#Sample & Re-sampling Time(s)\\
\hline
No re-sampling 
& $\texttt{ORG}$     & 0.587$\pm$0.001 & 0.721$\pm$0.000 & 0.632$\pm$0.011 & 0.663$\pm$0.026 & 0.803$\pm$0.001 & 170885 & - - -\\
\hline
\multirow{7}*{Under-sampling}
& $\texttt{RandUnder}$  & 0.022$\pm$0.008 & 0.068$\pm$0.000 & 0.011$\pm$0.001 & 0.013$\pm$0.000 & 0.511$\pm$0.096 & 632 & 0.07\\
& $\texttt{NearMiss}$   & 0.003$\pm$0.003 & 0.010$\pm$0.009 & 0.002$\pm$0.000 & 0.002$\pm$0.001 & 0.050$\pm$0.000 & 632 & 2.06\\
& $\texttt{Clean}$      & 0.741$\pm$0.018 & 0.697$\pm$0.010 & 0.727$\pm$0.029 & 0.698$\pm$0.032 & 0.810$\pm$0.003 & 170,680 & 428.88\\
& $\texttt{ENN}$        & 0.692$\pm$0.036 & 0.668$\pm$0.013 & 0.637$\pm$0.021 & 0.644$\pm$0.026 & 0.799$\pm$0.004 & 170,779 & 423.86\\
& $\texttt{TomekLink}$  & 0.699$\pm$0.050 & 0.650$\pm$0.031 & 0.671$\pm$0.018 & 0.659$\pm$0.027 & 0.814$\pm$0.007 & 170,865 & 270.09\\
& $\texttt{AllKNN}$     & 0.692$\pm$0.041 & 0.668$\pm$0.012 & 0.652$\pm$0.023 & 0.652$\pm$0.015 & 0.808$\pm$0.002 & 170,765 & 1066.48\\
& $\texttt{OSS}$        & 0.711$\pm$0.056 & 0.650$\pm$0.029 & 0.671$\pm$0.025 & 0.666$\pm$0.009 & 0.825$\pm$0.016 & 163,863 & 240.95\\
\hline
\multirow{4}*{Over-sampling}
& $\texttt{RandOver}$   & 0.052$\pm$0.000 & 0.532$\pm$0.000 & 0.051$\pm$0.001 & 0.561$\pm$0.010 & 0.706$\pm$0.013 & 341,138 & 0.14\\
& $\texttt{SMOTE}$      & 0.046$\pm$0.001 & 0.362$\pm$0.005 & 0.093$\pm$0.002 & 0.087$\pm$0.005 & 0.672$\pm$0.026 & 341,138 & 1.23\\
& $\texttt{ADASYN}$     & 0.017$\pm$0.001 & 0.360$\pm$0.004 & 0.031$\pm$0.001 & 0.035$\pm$0.007 & 0.496$\pm$0.081 & 341,141 & 1.87\\
& $\texttt{BorderSMOTE}$& 0.067$\pm$0.006 & 0.579$\pm$0.010 & 0.145$\pm$0.003 & 0.126$\pm$0.011 & 0.242$\pm$0.020 & 341,138 & 1.89\\
\hline
\multirow{2}*{Hybrid-sampling}
& $\texttt{SMOTEENN}$   & 0.045$\pm$0.001 & 0.329$\pm$0.006 & 0.084$\pm$0.004 & 0.074$\pm$0.012 & 0.665$\pm$0.017 & 340,831 & 478.36\\
& $\texttt{SMOTETomek}$ & 0.046$\pm$0.001 & 0.362$\pm$0.004 & 0.094$\pm$0.004 & 0.094$\pm$0.004 & 0.682$\pm$0.033 & 341,138 & 293.75\\
\hline
\multirow{1}*{Under-sampling + Ensemble}
& $\texttt{SPE}_{10}$   & \textbf{0.755}$\pm$0.003 & \textbf{0.767}$\pm$0.001 & \textbf{0.783}$\pm$0.015 & \textbf{0.808}$\pm$0.004 & \textbf{0.849}$\pm$0.002 & 632$\times$10 & 0.116$\times10$\\
\hline
\end{tabular}
\end{table*}

\subsubsection{Setup Details}
For each real-world task, we use 60\% of the full data as the training set $\mathcal{D}$ and 20\% as the validation set $\mathcal{D}_{dev}$ (some classifiers like GBDT need validation set for early stopping), the rest 20\% is then used as the test set $\mathcal{D}_{test}$.
All results in this section were evaluated on the test set in order to test the classifier's generalized performance.

\subsubsection{Results on Real-world Datasets}
We first extend the previous experiment on synthetic data to realistic datasets that we mentioned above. Table \ref{result-real} lists the experimental results of applying 6 different imbalance learning approaches (i.e., \texttt{RandUnder}, \texttt{Clean}, \texttt{SMOTE}, \texttt{Easy}, \texttt{Cascade}, and \texttt{SPE}) combine with 5 different canonical classification algorithms (i.e., KNN, DT, MLP, AdaBoost$_{10}$, and GBDT$_{10}$) on 5 real-world classification tasks\footnote{Due to space limitation, Table 5 only lists some most representative results. See
\texttt{github.com/ZhiningLiu1998/self-paced-ensemble}
for additional experimental results on more datasets and classifiers.}. The performance was evaluated by 4 criterions (i.e., AUCPRC, F1-score, G-mean, and MCC) on the test set. For reduce the effect of randomness, we show the mean and standard deviation of 10 independent runs:

\begin{itemize}[leftmargin=0.14in]
    \item \texttt{SPE} demonstrates the best performance on all tested real-world tasks using 5 classifiers over 4 evaluation criteria.
    \item \texttt{Clean} + MLP performs poorly on Credit Fraud task since \texttt{Clean} only cleans up noises and does not guarantee a balanced dataset. As described above, the batch training method will fail when the class distribution is skewed.
    \item \texttt{RandUnder} and $\texttt{Easy}_{10}$ use randomized under-sampling to get a small majority subset for training. They suffer from severe information loss and high potential variance when applying on highly imbalanced dataset.
\end{itemize}

Some results of \texttt{Clean} and \texttt{SMOTE} are missing in Table \ref{result-real} due to lack of appropriate distance metric and unacceptable computational cost. Take the KDDCUP (DOS vs. PRB) dataset as an example, from our experiment, \texttt{Clean} needs more than 8 hours to calculate the distance between each data sample. Similarly, \texttt{SMOTE} generates millions of synthetic samples that further enlarge the scale of the training set.

\subsection{Extensive Experiments on Real-world Datasets}
We further introduce some other widely used re-sampling and ensemble-based imbalance learning methods for a more comprehensive comparison. By showing supplementary information, e.g., the number of samples used for training and the processing time, we demonstrate the efficiency of different methods on real-world highly imbalanced tasks.

\subsubsection{Comparison with Re-sampling Methods}
9 more re-sampling based imbalance learning methods were introduced, including 5 under-sampling methods, 3 over-sampling methods and 2 hybrid-sampling methods (see Table \ref{result-resample}):
\begin{itemize}[leftmargin=0.14in, label=-]
    \item \texttt{NearMiss}~\cite{mani2003nearmiss} selects $|\mathcal{P}|$ samples from the majority class for which the average distance of the k nearest samples of the minority class is the smallest.
    \item \texttt{ENN} ({\em Edited Nearest Neighbor})~\cite{wilson1972enn} aims to remove noisy samples from the majority class for which their class differs from the one of their nearest-neighbors. 
    \item \texttt{TomekLink}~\cite{tomek1976tomeklink} removes majority samples by detecting ``TomekLinks''. A TomekLink exists if two samples of different class are the nearest neighbors of each other.
    \item \texttt{AllKNN}~\cite{tomek1976allknn} extends \texttt{ENN} by repeating the algorithm multiple times, the number of neighbors of the internal nearest neighbors algorithm is increased at each iteration.
    \item \texttt{OSS} ({\em One Side Sampling})~\cite{kubat1997oss} makes use of \texttt{TomekLink} by running it multiple times to iteratively decide if a sample should be kept in a dataset or not.
    \item \texttt{RandOver} (Random Over-sampling) randomly repeats some minority samples to balance the class distribution.
    \item \texttt{ADASYN} ({\em ADAptive SYNthetic over-sampling})~\cite{he2008adasyn} focuses on generating samples next to the original samples which are wrongly classified using a k-nearest neighbors classifier.
    \item \texttt{BorderSMOTE} ({\em Borderline Synthetic Minority Over-sampling TechniquE})~\cite{han2005borderline-smote} offers a variant of the \texttt{SMOTE} algorithm, where only the borderline examples could be seeds for over-sampling.
    \item \texttt{SMOTEENN} ({\em SMOTE with Edited Nearest Neighbours cleaning})~\cite{batista2004smoteenn} utilizes \texttt{ENN} as the cleaning method after applying \texttt{SMOTE} over-sampling to obtain a cleaner space.
    \item \texttt{SMOTETomek} ({\em SMOTE with Tomek links cleaning})~\cite{batista2003smotetomek} uses \texttt{TomekLink} instead of \texttt{ENN} as the cleaning method.
\end{itemize}

As mentioned before, running some of these re-sampling methods on large-scale datasets can be extremely slow. 
It is also hard to define an appropriate distance metric on a dataset with non-numerical features. 
With these considerations, we apply all methods on the Credit Fraud dataset.
This dataset has 284,807 samples, and only contains normalized numerical features, which enables all distance-based re-sampling methods to achieve their maximum potential. 
Thus we can fairly compare the pros and cons of different methods.

We use 5 different classifiers, i.e., Logistic Regression (LR), KNN, DT, AdaBoost$_{10}$, and GBDT$_{10}$, to collaborate with: 
$\texttt{ORG}$ which refers to train classifier over the original training set with no re-sampling, 12 re-sampling methods which refer to train classifier on the re-sampled training set, and \texttt{SPE} which refers to use our proposed method to get an ensemble of the given classifier.
We also list the number of examples that used for training and the time it takes to perform re-sampling for each method. All aforementioned re-sampling methods were implemented using imbalanced-learn Python package 0.4.3~\cite{guillaume2017imblearn} with Python 3.7.1, and run on an Intel Core i7-7700K CPU with 16 GB RAM.
Experimental results were shown in Table \ref{result-resample}:
\begin{itemize}[leftmargin=0.14in]
    \item \texttt{SPE} significantly boosts the performance of various canonical classification algorithms on highly imbalanced dataset. Comparing with other re-sampling methods, it only requires very little training data and short processing time to achieve such effects.
    \item Most methods can only obtain reasonable results (better than \texttt{ORG}) when cooperating with specific classifiers. For instance, \texttt{TomekLink} works well with LR, DT, and GBDT but fails to boost the performance of KNN and AdaBoost. The reason behind is that they perform model-agnostic re-sampling without considering classifier's capacity.
    \item On a dataset with such high imbalance ratio (IR=578.88:1), the minority class is often poorly represented and lacks a clear structure. Therefore, straightforward application of re-sampling, especially over-sampling that rely on relations between minority objects can actually deteriorate the classification performance, e.g., advanced over-sampling method \texttt{SMOTE} perform even worse than \texttt{RandOver} and \texttt{ORG}.
\end{itemize}

\begin{table*}[h]
\scriptsize
\renewcommand{\arraystretch}{1.0}
\caption{Generalized performance of 6 ensemble methods with different amount of base classifiers.}
\label{result-ensemble}
\centering
\begin{tabular}{|bc|tc|tctctctctc|tc|}
\hline
\rowstyle{\bfseries}
\# Base Classifiers & Metric & $\texttt{RUSBoost}_{n}$ & $\texttt{SMOTEBoost}_{n}$ & $\texttt{UnderBagging}_{n}$ & $\texttt{SMOTEBagging}_{n}$ & $\texttt{Cascade}_{n}$ & $\texttt{SPE}_{n}$\\
\hline
\multirow{5}*{$n=10$}
 & AUCPRC & 0.424$\pm$0.061 & 0.762$\pm$0.011 & 0.355$\pm$0.049 & 0.782$\pm$0.007 & 0.610$\pm$0.051 & {\bf 0.783}$\pm$0.015\\
 & F1 & 0.622$\pm$0.055 & {\bf 0.842}$\pm$0.012 & 0.555$\pm$0.053 & 0.818$\pm$0.002 & 0.757$\pm$0.031 & 0.832$\pm$0.018\\
 & GM & 0.637$\pm$0.045 & {\bf 0.847}$\pm$0.014 & 0.577$\pm$0.044 & 0.819$\pm$0.002 & 0.760$\pm$0.031 & 0.835$\pm$0.018\\
 & MCC & 0.189$\pm$0.016 & 0.822$\pm$0.018 & 0.576$\pm$0.044 & 0.819$\pm$0.002 & 0.759$\pm$0.031 & {\bf 0.835}$\pm$0.018\\
 \cline{2-8}
 & \# Sample & 6,320 & 1,723,295 & 6,320 & 1,876,204 & 6,320 & 6,320\\
\hline
\multirow{5}*{$n=20$}
 & AUCPRC & 0.550$\pm$0.032 & 0.783$\pm$0.005 & 0.519$\pm$0.125 & 0.804$\pm$0.013 & 0.673$\pm$0.008 & {\bf 0.811}$\pm$0.005\\
 & F1 & 0.722$\pm$0.021 & 0.840$\pm$0.009 & 0.678$\pm$0.088 & 0.833$\pm$0.005 & 0.779$\pm$0.012 & {\bf 0.856}$\pm$0.008\\
 & GM & 0.725$\pm$0.019 & 0.844$\pm$0.009 & 0.685$\pm$0.078 & 0.837$\pm$0.005 & 0.785$\pm$0.010 & {\bf 0.858}$\pm$0.010\\
 & MCC & 0.202$\pm$0.006 & 0.833$\pm$0.005 & 0.685$\pm$0.078 & 0.837$\pm$0.005 & 0.784$\pm$0.010 & {\bf 0.857}$\pm$0.010\\
 \cline{2-8}
 & \# Sample & 12,640 & 3,478,690 & 12,640 & 3,752,408 & 12,640 & 12,640\\
\hline
\multirow{5}*{$n=50$}
 & AUCPRC & 0.714$\pm$0.025 & 0.786$\pm$0.009 & 0.676$\pm$0.022 & 0.818$\pm$0.004 & 0.696$\pm$0.028 & {\bf 0.822}$\pm$0.006\\
 & F1 & 0.784$\pm$0.010 & 0.825$\pm$0.010 & 0.773$\pm$0.006 & 0.839$\pm$0.009 & 0.776$\pm$0.009 & {\bf 0.865}$\pm$0.012\\
 & GM & 0.784$\pm$0.010 & 0.830$\pm$0.010 & 0.774$\pm$0.006 & 0.843$\pm$0.008 & 0.785$\pm$0.011 & {\bf 0.868}$\pm$0.012\\
 & MCC & 0.297$\pm$0.015 & 0.794$\pm$0.007 & 0.774$\pm$0.006 & 0.842$\pm$0.008 & 0.784$\pm$0.011 & {\bf 0.868}$\pm$0.012\\
 \cline{2-8}
 & \# Sample & 31,600 & 8,937,475 & 31,600 & 9,381,020 & 31,600 & 31,600\\
\hline
\end{tabular}
\end{table*}

\begin{table*}[h]
\scriptsize
\renewcommand{\arraystretch}{1}
\caption{Performance (AUCPRC) of 6 ensemble methods with missing values.}
\label{result-missing}
\centering
\begin{tabular}{|bc|tctctctctc|tc|}
\hline
\rowstyle{\bfseries}
Missing Ratio & $\texttt{RUSBoost}_{10}$ & $\texttt{SMOTEBoost}_{10}$ & $\texttt{UnderBagging}_{10}$ & $\texttt{SMOTEBagging}_{10}$ & $\texttt{Cascade}_{10}$ & $\texttt{SPE}_{10}$\\
\hline
 0\%  & 0.424$\pm$0.061 & 0.762$\pm$0.011 & 0.355$\pm$0.049 & 0.782$\pm$0.007 & 0.610$\pm$0.051 & {\bf 0.783}$\pm$0.015\\
 25\% & 0.277$\pm$0.043 & 0.652$\pm$0.042 & 0.258$\pm$0.053 & 0.684$\pm$0.019 & 0.513$\pm$0.043 & {\bf 0.699}$\pm$0.026\\
 50\% & 0.206$\pm$0.025 & 0.529$\pm$0.015 & 0.161$\pm$0.013 & 0.503$\pm$0.020 & 0.442$\pm$0.035 & {\bf 0.577}$\pm$0.016\\
 75\% & 0.084$\pm$0.015 & 0.267$\pm$0.019 & 0.046$\pm$0.006 & 0.185$\pm$0.028 & 0.234$\pm$0.023 & {\bf 0.374}$\pm$0.028\\
\hline
\end{tabular}
\end{table*}

\subsubsection{Comparison with Ensemble Methods} 
In this experiment, we introduce four other ensemble based imbalance learning approaches for comparison:
\begin{itemize}[leftmargin=0.14in, label=-]
    \item \texttt{RUSBoost}
    \cite{seiffert2010rusboost}, which applies \texttt{RandUnder} within each iteration of Adaptive Boosting (AdaBoost) pipeline.
    \item \texttt{SMOTEBoost}
    \cite{chawla2003smoteboost}, which applies \texttt{SMOTE} to generate $|\mathcal{P}|$ new synthetic minority samples within each iteration of AdaBoost pipeline.
    \item $\texttt{UnderBagging}$ \cite{barandela2003underbagging} which applies \texttt{RandUnder} to get each bag for Bagging \cite{breiman1996bagging}. 
    Note that the only difference between $\texttt{UnderBagging}$ and \texttt{Easy} is that \texttt{Easy} use AdaBoost as its base classifier.
    \item \texttt{SMOTEBagging}
    \cite{wang2009smotebagging}, which applies \texttt{SMOTE} to get each bag for Bagging \cite{breiman1996bagging}, where each bag's sample quantity varies.
\end{itemize}

Our proposed method was then compared with 4 above methods and the \texttt{Cascade} that we used before.
They were considered as the most effective and widely-used imbalance learning methods in very recent reviews \cite{albert02018experiment,haixiang2017overview}.
Considering most of the previous approaches were proposed in combination with C4.5 \cite{barandela2003underbagging,wang2009smotebagging,seiffert2010rusboost}, for a fair comparison, we also use the C4.5 classifier as the base model in this experiment.
\texttt{Easy} was not included here since it is equivalent to $\texttt{UnderBagging}$ when cooperating with C4.5 classifier.

Because the number of base models significantly influences the performance of ensemble methods,
we test each method with 10, 20 and 50 base models in its ensemble.
We must note that such comparison is not totally fair since over-sampling methods need far more data and resources to train each base model.
In consideration of computational cost ($\texttt{SMOTEBoost}$ and $\texttt{SMOTEBagging}$ generate a huge amount of synthetic samples on large-scale highly-imbalanced dataset, see Table \ref{result-ensemble}), all ensemble methods were applied on the Credit Fraud dataset with AUCPRC, F1-score, G-mean, MCC for evaluation. For each method, we also list the total number of data samples (\# Samples.) that used for training all base models in the ensemble. 
Table \ref{result-ensemble} shows the experimental results of 5 ensemble methods and our proposed method:
\begin{itemize}[leftmargin=0.14in]
    \item Comparing with other 3 under-sampling based ensemble methods,
    \texttt{SPE} uses the same amount of training data but significantly outperforms them over 4 evaluation criteria.
    \item Comparing with 2 over-sampling based ensemble methods,
    \texttt{SPE} demonstrates competitive performance using far less (around 1/300) training data.
    \item Over-sampling based methods are woefully sample-inefficient. They generate a substantial number of synthetic samples under high imbalance ratio. As a result, they further enlarge the scale of training set thus need far more computing resources to train each base model. Higher imbalance ratio and larger dataset can make this situation even worse.
\end{itemize}

\begin{figure}[h]
    \centering
    \subfigure[Credit Fraud]{
        \includegraphics[width=0.46\linewidth]{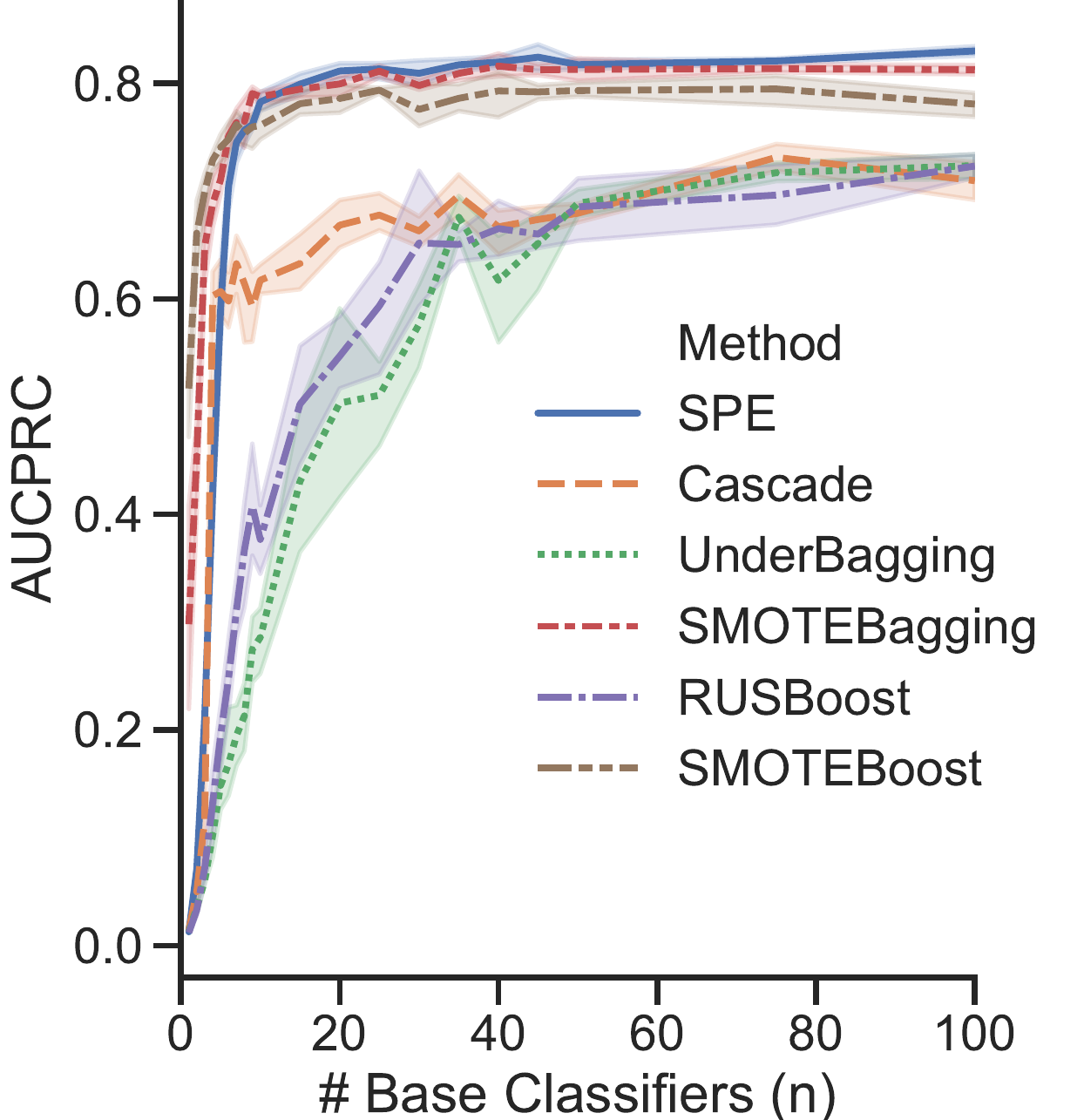}
        \label{fig:ensemble-credit}
    }
    \subfigure[Payment Simulation]{
        \includegraphics[width=0.46\linewidth]{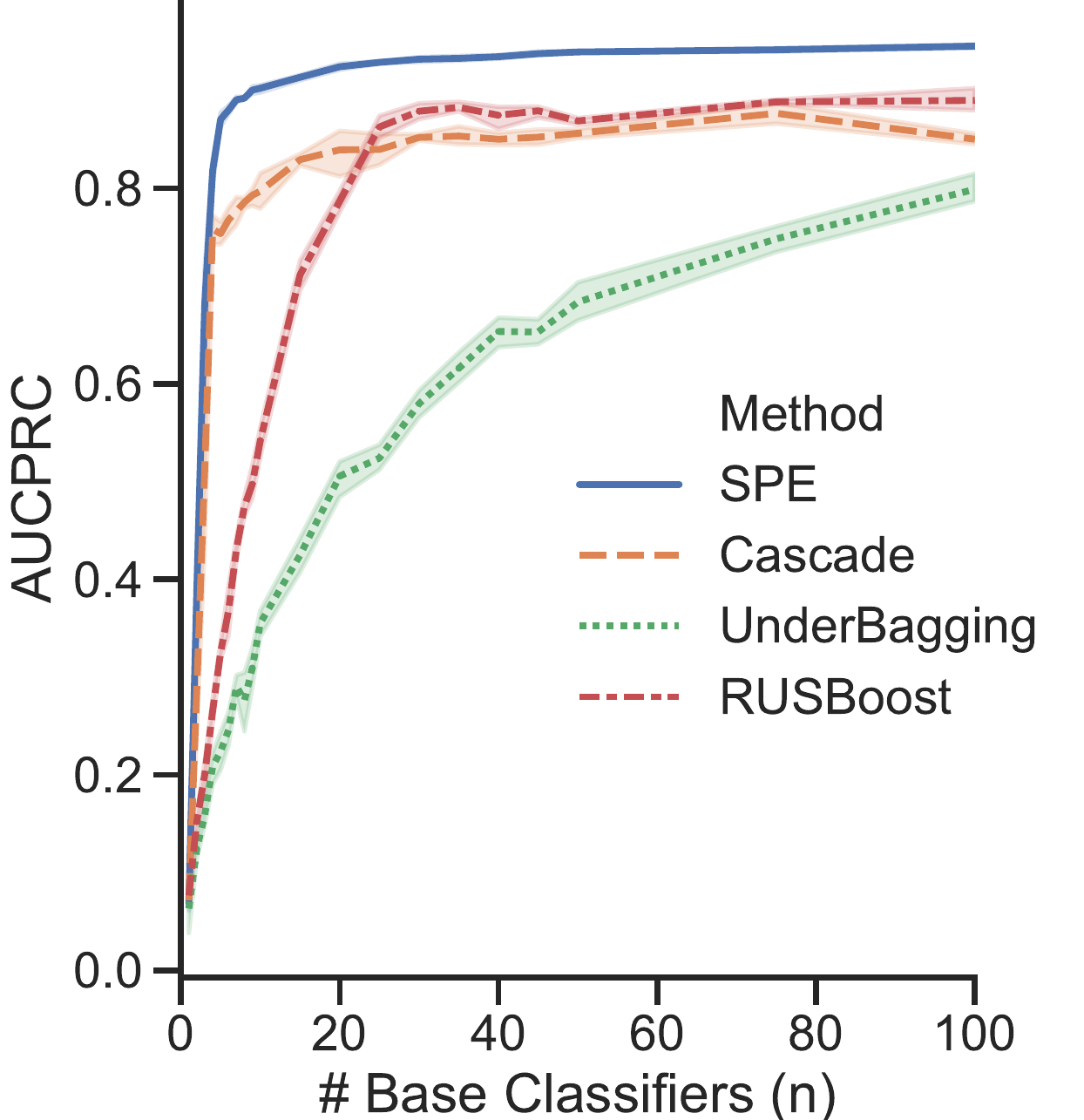}
        \label{fig:ensemble-paysim}
    }
    \caption{
    Generalized performance of ensemble methods on two real-world tasks with the number of base classifiers ($n$) ranging from 1 to 100. Each curve shows the results of 10 independent runs. Notice that the results of $\texttt{SMOTEBoost}$ and $\texttt{SMOTEBagging}$ are missing on Payment Simulation task due to lack of appropriate distance metric and large computational cost.
    }
    \label{fig:ensemble}
\end{figure}

We conduct more detailed experiments on Credit Fraud and Payment Simulation datasets, as shown in Fig. \ref{fig:ensemble}. 
We can see that although \texttt{SPE} uses little data for training, it can still obtain a desirable result which is even better than over-sampling based methods. Moreover, on both tasks \texttt{SPE} shows consistent performance in multiple independent runs. Compared to \texttt{SPE}, other methods are less stable and have greater randomness.

\subsubsection{Robustness under Missing Values} 
Finally, we test the robustness of different ensemble methods when there are missing values in the dataset. It is also a common problem that widely existing in real-world applications. To simulate the situation of missing values, we randomly select values from all features in both training and test datasets, then replace them with meaningless 0. We tested all methods on the Credit Fraud dataset, where 0\% / 25\% / 50\% / 75\% values are missing. 
Results were reported in Table \ref{result-missing}. We can observe that \texttt{SPE} demonstrates robust performance under different level of missing, while other methods performing poorly when the missing ratio is high.
We also notice that tested methods show different sensitivity to missing values. 
For an example, $\texttt{SMOTEBagging}$ obtains results better than $\texttt{SMOTEBoost}$ on the original dataset, but this situation is reversed when the missing ratio is greater than 50\%.

\subsubsection{Sensitivity to Hyper-parameters}
\begin{figure}[h]
    \centering
    \subfigure[Credit Fraud]{
        \includegraphics[width=0.4\linewidth]{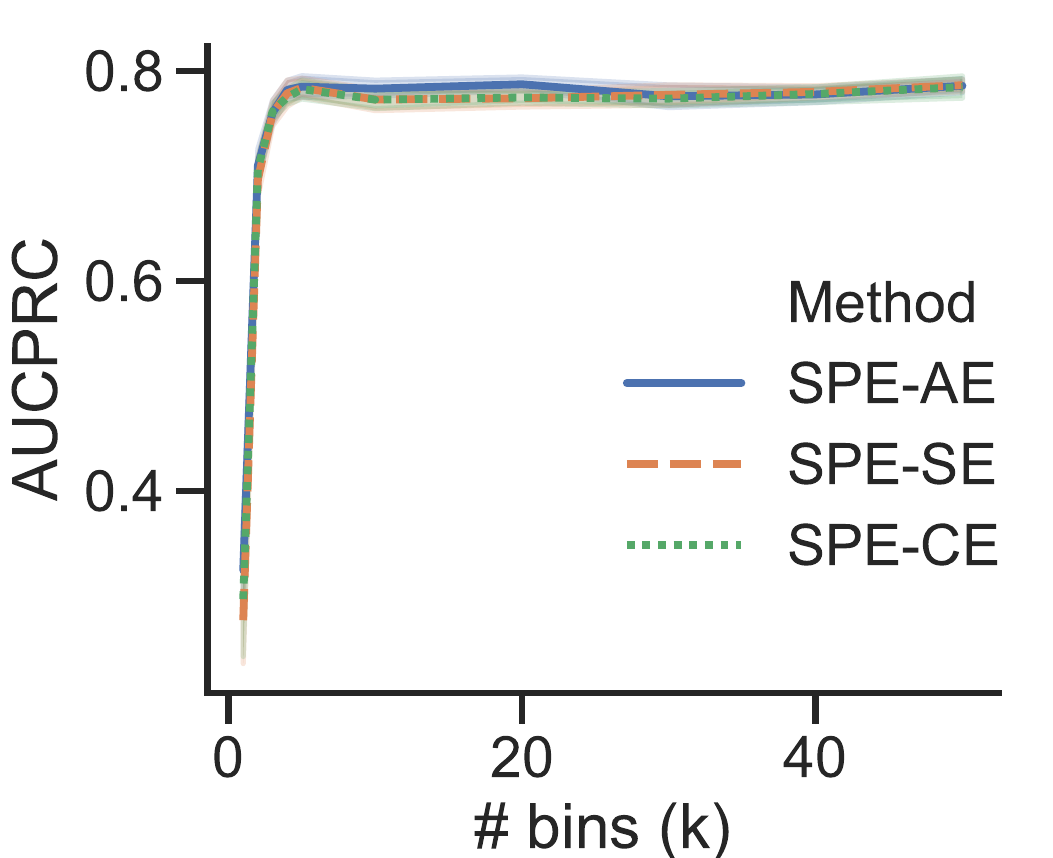}
        \label{fig:sens-credit}
    }
    \subfigure[Payment Simulation]{
        \includegraphics[width=0.4\linewidth]{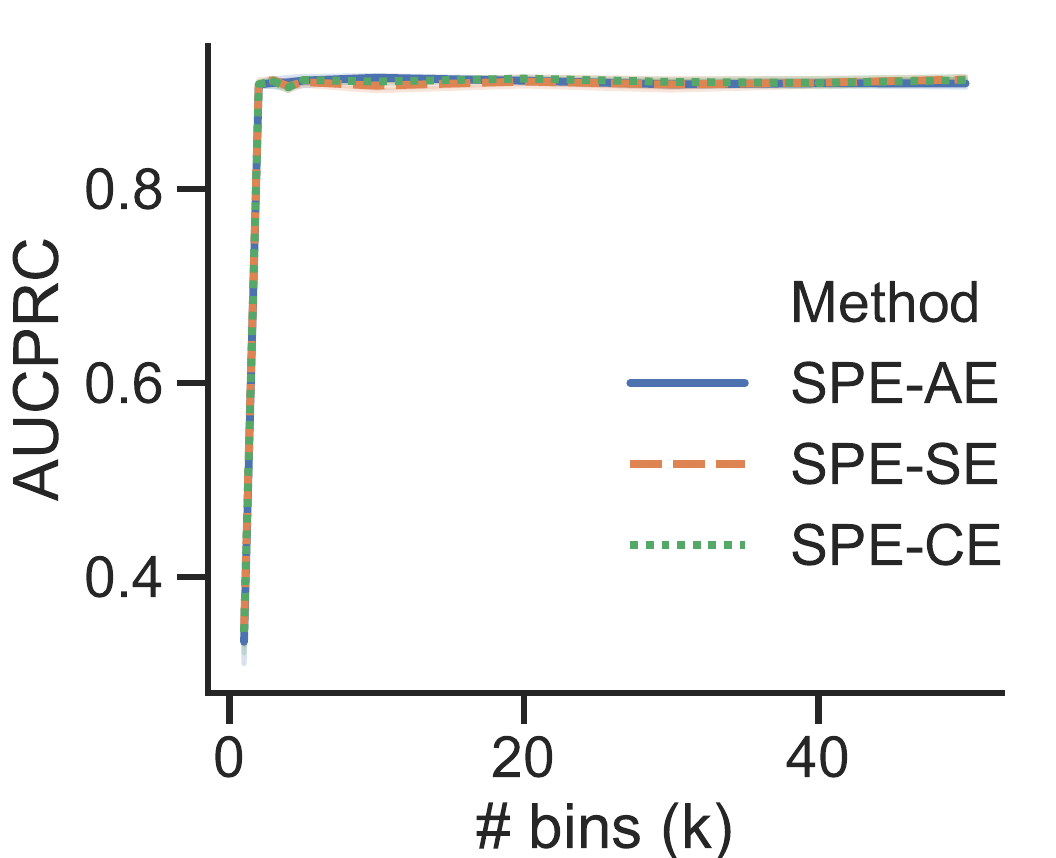}
        \label{fig:sens-paysim}
    }
    \caption{
    Performance (mean of 10 independent runs) of $\texttt{SPE}_{10}$ on two real-world tasks using different number of bins ($k$) and hardness function ($\mathcal{H}$).
    }
    \label{fig:sensitivity}
\end{figure}

\texttt{SPE} has 3 key hyper-parameters: number of base classifiers $n$, number of bins $k$ and hardness function $\mathcal{H}$. In previous discussion we demonstrate the influence of the number of base classifiers ($n$). Now we conduct experiment to verify the impact of the number of bins ($k$) and different choices of hardness function ($\mathcal{H}$). Specifically, we test $\texttt{SPE}_{10}$ on two real-world tasks with $k$ ranging from 1 to 50, in cooperation with 3 different hardness functions. They are Absolute Error (AE), Squared Error (SE) and Cross Entropy (CE), where: 
\begin{enumerate}
    \item $\mathcal{H}_{AE}(x,y,F) = |F(x)-y|$
    \item $\mathcal{H}_{SE}(x,y,F) = (F(x)-y)^2$
    \item $\mathcal{H}_{CE}(x,y,F) = -ylog(F(x))-(1-y)log(1-F(x))$
\end{enumerate}
The results in Fig. \ref{fig:sensitivity} show that our method is robust to different selection of $k$ and $\mathcal{H}$. Note that $k$ determines how detailed our hardness distribution approximation is, thus setting a small $k$, e.g., $k<10$, may lead to poor performance. 

\section{Related work}

Imbalanced data classification has been a fundamental problem in machine learning \cite{he2008overview,he2013overview}. Many research works have been proposed to solve such problem. This research field is also known as {\em Imbalance Learning}.
Recently, Guo {\em et al.} provided a systematic review of existing methods and real-world applications in the field of imbalance learning \cite{haixiang2017overview}. 

Most of proposed works employed distance-based methods to obtain re-sampled data for training canonical classifiers \cite{laurikkala2001ncr,tomek1976tomeklink,chawla2002smote,he2008adasyn}.
Based on them, many works combine re-sampling with ensemble learning \cite{seiffert2010rusboost,chawla2003smoteboost,barandela2003underbagging,wang2009smotebagging}. Such strategies have proven to be very effective~\cite{albert02018experiment}.
Distance-based methods have several deficiencies. 
First, it is hard to define distance on a real-world dataset, especially when it contains categorical features or missing values. 
Second, the cost of computing distances between each samples can be huge when applying on large-scale datasets.
Even though the distance-based methods have been successfully used for re-sampling, they do not guarantee desirable performance for different classifiers due to their model-agnostic designs. 

Some other methods try to assigning different weights to samples rather than re-sampling the whole dataset \cite{elkan2001cost-sensitive,liu2006cost-sensitive-imbalance}. They require assistance from domain experts and may fail when cooperating with batch training methods (e.g. neural network). We prefer not to include such methods in this paper because previous experiments \cite{liu2006cost-sensitive-imbalance} have shown that setting arbitrary costs without domain knowledge do not allow them to achieve their maximum potential.

There are some works in other domains (e.g. Active Learning \cite{settles2009active-learning}, Self-paced Learning \cite{kumar2010spl}) that adopt the idea of selecting ``informative'' samples but focus on completely different problems. Specifically, an active learner interactively queries the user to obtain the labels of new data points, while a self-paced learning algorithm tries to present the training data in a meaningful order that facilitates learning. However, they perform the sampling without considering the overall data distribution, thus their fine-tuning process can be easily disturbed when the training set is imbalanced. In comparison, \texttt{SPE} applies under-sampling + ensemble strategy to balance the dataset, making it applicable to any canonical classifier. By considering the dynamic hardness distribution over the whole dataset, \texttt{SPE} performs adaptive and robust under-sampling rather than blindly selecting ``informative'' data samples.

To summarize, traditional distance-based re-sampling methods ignore the difference of model capacity, thus may lead to poor performance when cooperating with specific classifiers. They also require additional computation to calculate distances between samples, making them computationally inefficient, especially on large datasets. Moreover, it is often difficult to determine a clear distance metric in practice, as real-world datasets may contain categorical features and missing values. Most ensemble-based methods integrate such distance-based re-sampling into their pipelines, thus are still negatively affected by the above factors.
Comparing with existing works, \texttt{SPE} doesn't require any pre-defined distance metric or computation, making it easier to apply and more computationally efficient. By self-paced harmonizing the hardness distribution w.r.t the given classifier, \texttt{SPE} is adaptive to different models and robust to noises and missing values.
\section{Conclusions}

In this paper we have described the problem of {\em highly imbalanced}, {\em large-scale} and {\em noisy} data classification that widely exists in real-world applications. Under such a scenario, we have demonstrate that canonical machine learning / imbalance learning approaches suffer from unsatisfactory results and low computational efficiency.

{\em Self-paced Ensemble}, a novel learning framework for massive imbalance classification has been proposed in this paper.
We argue that all of the difficulties - high imbalance ratio, overlapping between classes, presence of noises - are critical for massive imbalance classification. 
Hence, we have introduced the concept of classification hardness distribution to integrate the information of these difficulties into our learning framework.
We conducted extensive experiments on a variety of challenging real-world tasks. Comparing with other methods, our framework has better performance, wider applicability, and higher computational efficiency.
Overall, we believe that we have provided a new paradigm of integrating task difficulties into the imbalance classification system. Various real-world applications can benefit from our framework. 

\bibliographystyle{./bibliography/IEEEtran}
\bibliography{./bibliography/IEEEabrv,./bibliography/IEEEexample}
\end{document}